\documentclass[sn-apa, iicol]{sn-jnl}

\jyear{2021}%

\theoremstyle{thmstyleone}%
\theoremstyle{thmstyletwo}%

\theoremstyle{thmstylethree}%
\usepackage{caption}
\usepackage{subcaption}

\usepackage{stmaryrd}
\usepackage{booktabs}

\begin{document}

\title[Article Title]{\begin{center} Fusion for Visual-Infrared Person ReID in Real-World Surveillance Using Corrupted Multimodal Data \\ - \end{center} First version submitted to IJCV}




\author{\fnm{Arthur} \sur{Josi}}\email{\{arthur.josi.1, mahdi.alehdaghi.1\}@ens.etsmtl.ca}

\author{\fnm{Mahdi} \sur{Alehdaghi}}

\author{\fnm{Rafael} \sur{M. O. Cruz}}\email{\{rafael.menelau-cruz, eric.granger\}@etsmtl.ca}

\author{\fnm{Eric} \sur{Granger}}

\affil{\centering \orgdiv{Laboratoire d’imagerie, de vision et d’intelligence artificielle (LIVIA)} \\ \orgdiv{ETS Montreal}, \country{Canada}}



\abstract{
Visible-infrared person re-identification (V-I ReID) seeks to match images of individuals captured over a distributed network of RGB and IR cameras. The task is challenging due to the significant differences between V and I modalities, especially under real-world conditions, where images are corrupted by, e.g,  blur, noise, and weather. Despite their practical relevance, deep learning (DL) models for multimodal V-I ReID remain far less investigated than for single and cross-modal V to I settings. Moreover, state-of-art V-I ReID models cannot leverage corrupted modality information to sustain a high level of accuracy. In this paper, we propose an efficient model for multimodal V-I ReID -- named Multimodal Middle Stream Fusion (MMSF) --  that preserves modality-specific knowledge for improved robustness to corrupted multimodal images. In addition, three state-of-art attention-based multimodal fusion models are adapted to address corrupted multimodal data in V-I ReID, allowing for dynamic balancing of the importance of each modality. The literature typically reports ReID performance using clean datasets, but more recently, evaluation protocols have been proposed to assess the robustness of ReID models under challenging real-world scenarios, using data with realistic corruptions. However, these protocols are limited to unimodal V settings. For realistic evaluation of multimodal (and cross-modal) V-I person ReID models, we propose new challenging corrupted datasets for scenarios where V and I cameras are co-located (CL) and not co-located (NCL). Finally, the benefits of our Masking and Local Multimodal Data Augmentation (ML-MDA) strategy are explored to improve the robustness of ReID models to multimodal corruption. Our experiments on clean and corrupted versions of the SYSU-MM01, RegDB, and ThermalWORLD datasets indicate the multimodal V-I ReID models that are more likely to perform well in real-world operational conditions. In particular, our ML-MDA is an important strategy for a V-I person ReID system to sustain high accuracy and robustness when processing corrupted multimodal images. The multimodal ReID models provide the best accuracy and complexity trade-off under both CL and NCL settings and compared to state-of-art unimodal ReID systems, except for the ThermalWORLD dataset due to its low-quality I. Our MMSF model outperforms every method under CL and NCL camera scenarios. GitHub code: \url{https://github.com/art2611/MREiD-UCD-CCD.git}.
}


\maketitle

\section{Introduction}

Real-world video monitoring and surveillance applications (e.g., recognizing individuals in airports, and vehicles in traffic) are challenging problems that rely on object detection \citep{zou2019object, zaidi2022survey}, tracking \citep{luo2021multiple}, classification \citep{sen2020supervised}, and re-identification (ReID) \citep{khan2019survey, PersonREID_outlook}. Person ReID aims to recognize individuals over a set of distributed non-overlapping cameras. State-of-art ReID systems based on, e.g., deep Siamese networks \citep{fu2021unsupervised, sharma2021person, somers2023body},  typically learn an embedding through various metric learning losses, which seeks to make image pairs with the same identity closer, and image pairs with different identities more distant in the embedding space. Despite the recent advances with deep learning (DL) models, person ReID remains a challenging task due to the non-rigid structure of the human body, the different viewpoints/poses with which a person can be observed, image corruption, and the variability of capture conditions (e.g., illumination, scale, contrast) \citep{bhuiyan2020pose, mekhazni2020unsupervised}. 

Visible-infrared (V-I) person ReID aims to recognize individuals of interest across a network of RGB and IR cameras. Unlike visible cameras, infrared ones allow night-time recognition. This has motivated research on cross-modal recognition, to provide methods for V-I person ReID from night-time to day-time, or vice-versa \citep{PersonREID_outlook}. In addition, a V-I person ReID approach has been proposed for a multimodal recognition \citep{RegDB}, where the I modality is used in conjunction with the V, improving accuracy due to its different data encoding and perception under low light conditions. In fact, a V-I ReID can allow training a single model remains accurate over diverse capture conditions. 
A V-I ReID model should however conserve modality specific-features instead of focusing mostly on modality-shared ones \citep{baltruvsaitis2018multimodal}, which is often absent, or not explicitly addressed by state-of-art approaches. Furthermore, RGB and IR cameras may be co-located (CL) or not co-located (NCL), and variation in camera configuration affects the spacial alignment of V-I images, which is likely influencing ReID (as it is known to impact other tasks) \citep{wang2021role, 9745968}. 

Artificially corrupted datasets \citep{hendrycks2019benchmarking,chen2021benchmarksDataset_C, michaelis2019benchmarking} are important for evaluating V-I person ReID models, yet public datasets are often collected in controlled environments that cannot cover the range of real-world scenarios \citep{poria2017review}. As highlighted by \cite{rahate2022multimodal}, there is a need to create multimodal real-world datasets that contain corrupted modalities. Apart from the recent approach using corrupted audio-visual data in emotion recognition \cite{hong2023watch}, the V-I ReID evaluation set proposed in our preliminary work \citep{our_preliminary} is, to our best knowledge, the only existing dataset for corrupted evaluation for visual multimodal learning. However, the dataset in \citep{our_preliminary} is only evaluated for a simple architecture, and does not consider the correlation in the corruption from one camera to another. For example, corruption due to weather conditions should similarly occur on a V-I pair from co-located V-I cameras. 

Neglecting to evaluate ReID models on corrupted data can result in large and unexpected performance gaps at deployment. To reduce this gap, one can attempt to restore corrupted input images during test time \citep{chang2020siamese}, at the expense pipeline complexity by restoring the data before proceeding to the main ReID task. Using more complex DL models has been shown to improve performance on corrupted image data in object detection \citep{michaelis2019benchmarking} and image classification \citep{xie2020self}. For instance, vision transformer models \citep{han2020survey} have been shown some robustness to image corruption \citep{hendrycks2020pretrained}. In particular, the TransReID model \cite{he2021transreid} provides state-of-art person ReID performance when facing corrupted data \citep{chen2021benchmarksDataset_C}. However, such complex models limit the potential for real-time ReID applications. Using more diverse training data can improve the robustness of deep ReID models to corrupted data \citep{xie2020self}, and does not increase the model's complexity at test time. Data augmentation \citep{shorten2019survey} also avoids the costs of data collection and annotation. 

This paper focuses on the following research questions. How can efficient V-I ReID models be developed considering CL or NCL scenarios? How can these V-I models be trained, thanks to augmented multimodal data, to provide better robustness to real-world image corruptions than state-of-art models like TransReID? 
In this paper, a cost-effective V-I ReID model named Multimodal Middle Stream Fusion (MMSF) is proposed to explicitly preserve and exploit both modality-specific and modality-shared knowledge, thereby improving robustness to corrupted images. In addition, three state-of-art attention-based models are adapted from the areas of sentiment analysis, emotion recognition, and action recognition for similarity matching, as needed for person ReID. Attention approaches are expected to address image corruptions through a dynamic feature selection, dealing with the varying availability of modality information. However, these models mainly focus on modality-shared features, eventually losing some capacity to  discriminate. 

Essential for the evaluation of both multimodal and cross-modal V-I person ReID models, corrupted V-I datasets are proposed for uncorrelated and correlated cases, named respectively uncorrelated corrupted dataset (UCD) and correlated corrupted dataset (CCD). These two sets allow for a robust evaluation of models based on 20 V and 19 I different corrupted conditions. Improving from our preliminary work, corruptions are correlated or not to suit NCL and CL camera configurations. In our experiments, we validate ReID models using clean and corrupted versions of the SYSU-MM01 \citep{SYSU} (NCL), RegDB \citep{RegDB} (CL), and ThermalWORLD \citep{ThermalWORLD} (CL) datasets. Our preliminary work in \citep{our_preliminary} introduced the Masking and Local Multimodal Data Augmentation (ML-MDA) strategy that improves the accuracy and robustness to strong corruptions using simple fusion architecture. The strategy is further assessed in this paper, and expected to train models leveraging the complementary knowledge among modalities while dynamically balancing the importance of individual modalities in final predictions.

\noindent \textbf{Main contributions:}

\noindent (1) A novel MMSF architecture is proposed for V-I ReID that allows preserving both modality-specific and -shared features. This aspect is shown to be essential for both CL and NCL settings but is not addressed most of the time. Additionally, three state-of-art attention-based models are adapted to similarity matching, and evaluated for V-I person ReID. These models are detailed in Section \ref{sec:models}.

\noindent (2) For realistic evaluation of V-I person ReID models, challenging UCD and CCD datasets are designed (see Section \ref{sec:UCD_CCD}). 

\noindent (3) The ML-MDA strategy presented Section \ref{sec:ML_MDA} is introduced for training DL models for V-I ReID multimodal that are robust to corruption.

\noindent (4) Our empirical results (see Section \ref{sec:results}) on clean and corrupted versions of the challenging SYSU-MM01, RegDB, and ThermalWORLD datasets provides insight about cost-effective DL models to adopt for V-I ReID, and their dependency on dataset properties and CL/NCL scenarios. Results also indicate that our V-I ReID models can outperform TransReID and related state-of-art models on clean and corrupted data in terms of accuracy and complexity.  


\section{Related Work}\label{sec:related_work}

\subsection{Multimodal fusion}

\noindent \textbf{Fusion approach and spatial alignment.} To better handle or analyze a given problem, not being restricted to a single source of information is usually a powerful strategy \citep{baltruvsaitis2018multimodal, wang2021survey}. As well-known approaches, one can think of late \citep{snoek2005late} or sensor  \citep{lohweg2010fuzzy} fusions. The former considers independent learning and feature extraction for each modality before making a decision.
Such fusions are easy to implement, as models can be trained independently and added to a system through minor adjustments. However, a model cannot learn the correlation between the modalities \citep{zhang2017learning}, like spatially related information. The latter (i.e., sensor fusion) stacks modalities together before any feature extraction, allowing inter-modality correlations to be mined and used by the model but considerably increasing the input dimension. Also, no spatial alignment may make modality correlations harder to find by the model \citep{wang2021role}. 

Intermediate or model-level fusion techniques consider fusing modalities during the feature extraction and before the decision layer \citep{baltruvsaitis2018multimodal}, increasing the semantic information contained in features before fusion and eventually making correlations easier to find. However, where spatial information continuously disappears through the network \citep{chen2018encoder}, it is unclear how much remains at each step and how it may impact a model. From experiments provided by \cite{wang2021role} on fusion location and data alignment, it is important to differentiate spatially aligned and unaligned data as models may have really distinct behaviors.  


\noindent \textbf{Model level fusion.} Model-level fusion considers fusing modality representations of a deep learning model somewhere in between the sensor representations and the feature vectors. Coordinated modality representation is seen by \cite{baltruvsaitis2018multimodal} as a challenging but promising fusion direction for model-level fusion approaches. Exchanging modality knowledge allows it and seems very practical as correlations may be mined by a model and as one modality may be more or less informative. However, they raise the models' lack of ability to conserve supplementary information and not only exploit complementary information. 

In practice, attention-based multimodal approaches allow modality knowledge exchange, as it is the case for the Multimodal Transfer Module (MMTM) proposed by \cite{mmtm2020joze}. The module refactors the channels of each modality regarding how the intra- and inter-modality channels correlate. Based on the MMTM concept and inspired by \cite{splitcnn2020zhang} that used the split operation to improve the dynamic channel selection, \cite{msaf2020su} presented the MSAF approach. The dynamic channel refactoring in such multimodal models may allow for fine-grained feature selection and limit corruption impact. Unlike previous approaches, modality attention \cite{gu2018hybrid}, later updated by \cite{ismail2020improving}, provides soft attention weights for each modality to balance modality importance in the final embedding based on their discriminating capabilities. Again, such attention sounds to be a great approach to tackling punctually corrupted data. However, those attention models do not explicitly work at conserving the modality-specific knowledge, missing the point raised by \cite{baltruvsaitis2018multimodal}.

Some transformers architectures tackle this aspect, conserving modality-specific knowledge through modality-specific streams and self-attention, and modality-shared knowledge thanks to modality-shared streams and cross-attention \citep{sun2021multimodal,lian2021ctnet,wei2020multi}. However, transformer architectures are known to be complex and heavy \cite{han2020survey}, which do not align with video-surveillance challenges, requiring close to real-time algorithms. 

\noindent \textbf{Multimodal person ReID.} Most approaches for person ReID \citep{PersonREID_outlook} focus on the unimodal (RGB) \citep{ristani2018features,luo2019bag} and cross-modal \citep{PersonREID_outlook,alehdaghi2022visible,zhang2022fmcnet} settings. Few only focused on combining multimodal information. For example, \cite{chen2019contour} used the contour information. \cite{bhuiyan2020pose} used pose information. However, for those approaches, the additional modality is built from the exploitation of the main modality, which would be similarly affected by image corruption and consequently not so helpful in this regard. 

Using another sensor to extract a supplementary modality allows to have a distinct encoding, likely differently affected by corruptions. For example, the infrared and near-infrared are shown to be beneficial for person ReID \citep{zheng2021robust, wang2022interact}, but leveraging the knowledge from three modalities might not be realistic for a real-world surveillance setting, asking for large models architectures. 

\cite{RegDB} represents the only approach where visible and infrared modalities only are integrated into a joint representation space. Infrared and visual features are concatenated, produced from independently trained CNNs, and used for pairwise matching at test time. This simple model attained an impressive performance on the RegDB dataset. However, RegDB data is captured with only one camera per modality, RGB-IR cameras are co-located with only a single tracklet of ten images per modality and individual, and except for their low resolutions, captured images present no specific corruptions. For these reasons, the RegDB dataset is less consistent with a real-world scenario. In fact, the development of person ReID models that are effective in uncontrolled real-world scenarios remains an open problem \citep{hendrycks2021many}. 

\subsection{Image corruption and augmentation strategies}    
    
Data augmentation (DA) consists in multiplying the available training dataset by punctually applying transformations on training images, like flips, rotations, and scaling \citep{ciregan2012multi}. This way, a model usually benefits from increased robustness to image variations and improved generalization performance. According to \cite{geirhos2018generalisation}, training a model on a given corruption is only sometimes helpful over other types of degradation. Yet, \cite{rusak2020simple} showed that a well-tuned DA can help the model to perform well over multiple types of image corruption through Gaussian and Speckle noise augmentation. \cite{hendrycks2019augmix} proposed the Augmix strategy, for which multiple variations of an image are obtained through randomly applied transformations, variations that get mixed together. Random Erasing occludes parts of the images punctually by replacing pixels with random values \citep{zhong2020random}. Previous strategies allow a large variety of augmented images, simulating eventually real-world data and hence inducing higher generalization performance. 

    \begin{figure*}
        \centering
        \includegraphics[width=\textwidth]{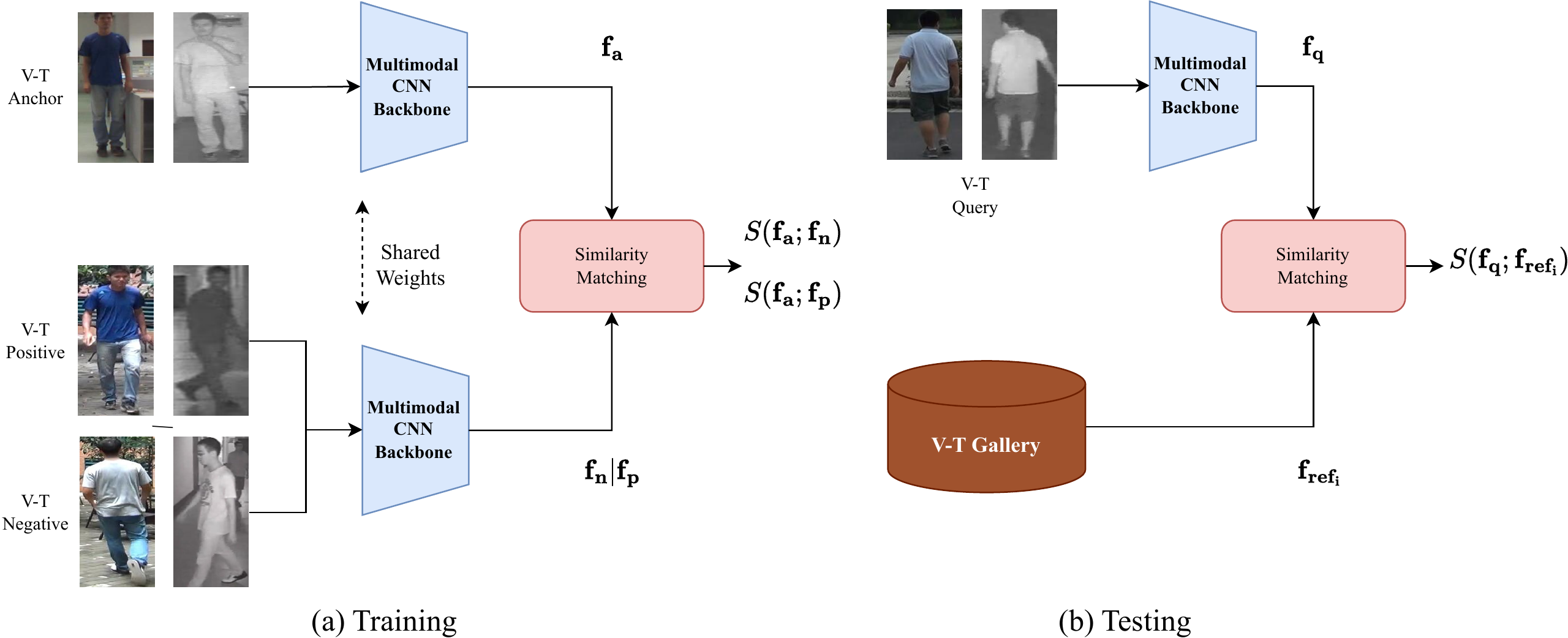}     
        \caption{Representation of the multimodal person ReID (a) learning while using the triplet loss and (b) inference.}
        \label{fig:global_multimodal_training_rpz}
    \end{figure*}
    
Focusing on person ReID, \cite{chen2021benchmarksDataset_C} proposed the CIL learning strategy to improve systems performance under corrupted data. Their strategy is partly based on two local DA methods -- self-patch mixing and soft random erasing. The former replaces some of the pixels in a patch with random values, while the latter superposes a randomly selected patch from an image at a random position on this same image. \cite{gong2021person} show interesting improvements through local and global grayscale patch DA on RGB images. However, the previous strategies are limited to single modality stream models, even though the latter shows how grayscale data may reinforce the visible modality features using DA. 

Multimodal data augmentation strategies have presented encouraging results for image-text emotion recognition \citep{xu2020mda} or vision-language representation learning \citep{hao2022mixgen}. Also, \cite{nakamura2022few} proposed a visible-thermal cross-domain DA for few shots object thermal detection, working at closing the domain gap by augmenting data through hetero modality objects added on the main modality images. However, to our best knowledge, our preliminary work \citep{our_preliminary} is the first to propose MDA with V-I person ReID applications through ML-MDA. Still, this MDA has only been investigated on a simple fusion model, which does not assure its generalization to more developed fusion architectures. Also, the evaluation is limited to corruptions set that do not consider eventual correlations between corruptions for NCL or CL cameras, which is tackled in this work.

\section{Multimodal Fusion for V-I ReID}\label{sec:models}

    The main objective of our study is to find how modalities should be fused to be robust to data corruption while conserving great performances on clean data. Hence, plural multimodal models are studied, all trained and evaluated following a pairwise matching scheme (Fig. \ref{fig:global_multimodal_training_rpz}). 
    
    From our preliminary work \citep{our_preliminary}, the learned concatenation model is now used as a baseline, referred to as Baseline C. Baseline S stands as our second baseline with the same architecture but an element-wise sum fusion of the feature vectors instead of a concatenation. 

    The selection of modality-shared and modality-specific features remains unclear in most models, whereas the importance of the conservation of both feature types has been highlighted by \cite{baltruvsaitis2018multimodal}. Hence, the multimodal middle stream fusion (MMSF) is proposed and first presented. Three attention-based models follow as attention should handle corruption well through a dynamic feature selection regarding each input. Still, the attention could also allow a modality corruption to degrade the hetero modality and require investigation. The three models are extracted from the literature and specially adapted to pairwise matching and, more precisely, to the person ReID task. 
    
    
    Fusion approaches are not restricted to a specific backbone, but ResNet-18 \citep{resnet_he2016deep} backbones are used for illustration purposes. Each model is optimized using the batch hard triplet loss \citep{TripletLoss} $\mathcal{L}_{\text{BH\_tri}}$, and cross-entropy with regularization via label smoothing \citep{szegedy2016rethinking} $\mathcal{L}_{\text{CE\_ls}}$. We follow the usual optimization process \citep{PersonREID_outlook}, except for the cross-entropy. Indeed, regularization via label smoothing is used by \cite{chen2021benchmarksDataset_C} is is better at addressing  corruption.

    \subsection{Multimodal middle stream fusion}\label{sec:MMSF}

        Assuring the conservation of the modality supplementary information, while taking advantage of the modality-shared information, we propose the Multimodal Middle Stream Fusion (MMSF). 
        
\begin{figure*}[!t]
            \begin{center}
            \includegraphics[width=0.8\linewidth, trim={0 0 4.7cm 0}]{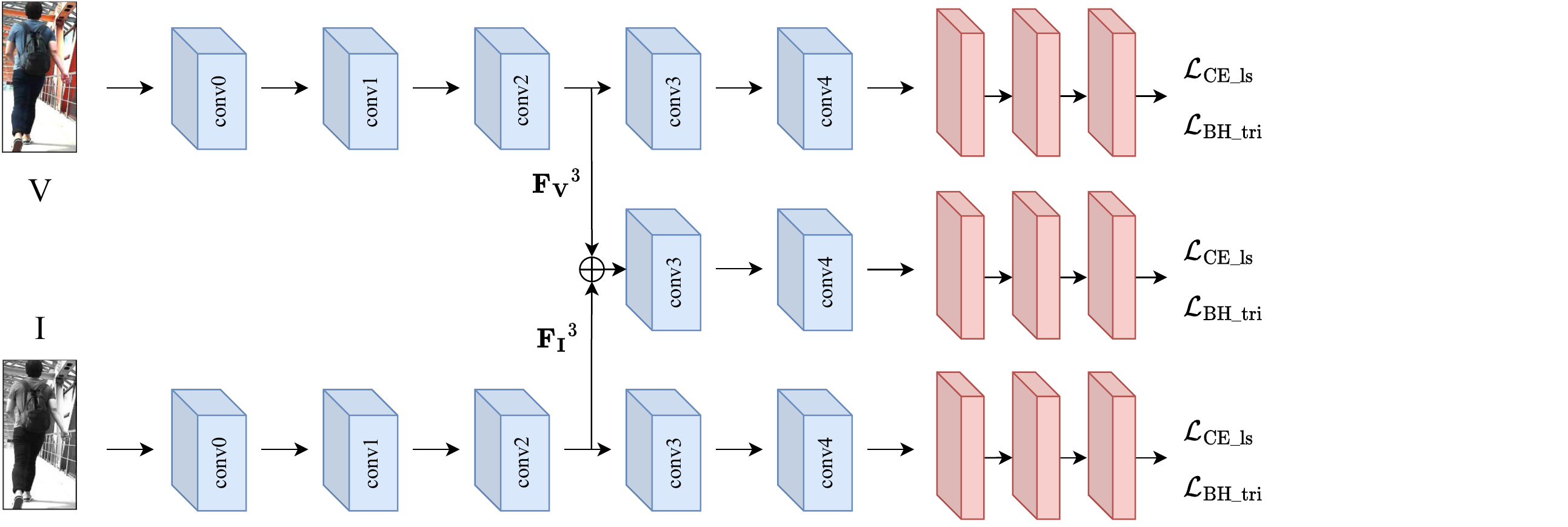}
            \end{center}
            \caption{Training architecture of the MMSF model while fusing the features in the middle stream for \textit{l}=3.}
            \label{fig:MMSF_training}
        \end{figure*}

        The model comprises two independent modality-specific CNN streams focused on the modality-specific information and a middle CNN stream that exploits the modality-shared information (Fig. \ref{fig:MMSF_training}). Each stream is independent and optimized through its specific loss functions, allowing it not to influence a stream representation from direct knowledge exchanges among streams. $\mathbf{F_{V}^\textit{l}}\in \mathbb{R}^{H \times W \times C}$ and $\mathbf{F_I^\textit{l}} \in \mathbb{R}^{H \times W \times C}$ are the visible and infrared feature maps before convolution blocks $l\in \mathbb{N}$. For a fusion before layer $l$, the middle stream takes $\mathbf{F_m}=\mathbf{F_{V}^\textit{l}} + \mathbf{F_I^\textit{l}}$ as input and pursues the feature extraction from this fused representation. Its middle stream size varies regarding $l$ value, being a partial backbone starting at layer $l$. 

        \subsection{Attention-based models}
        \subsubsection{Modality attention network}

        \begin{figure}[t]
            \begin{center}
            \includegraphics[width=1\linewidth]{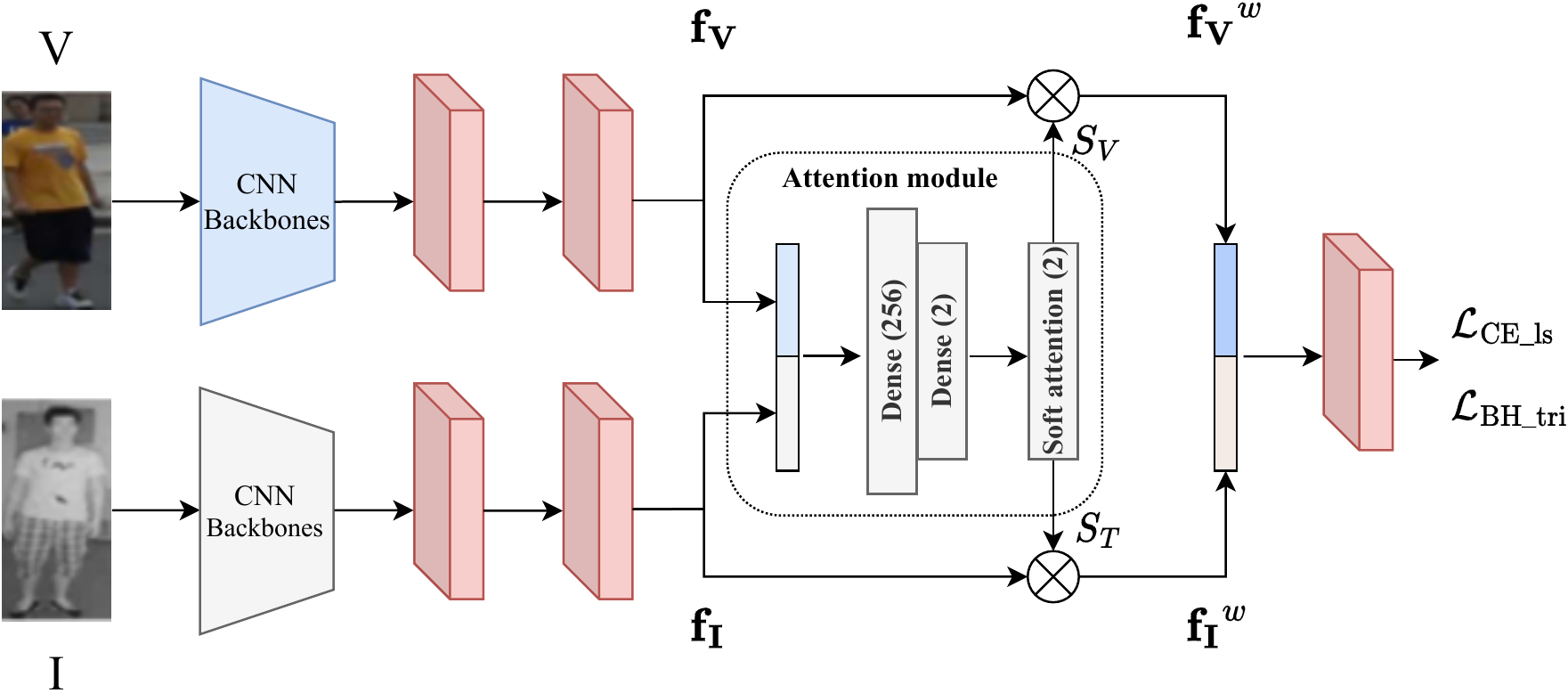}
            \end{center}
            \caption{Training architecture of the MAN model.}
            \label{fig:MAN}
        \end{figure}

        \begin{figure*}[t]
            \begin{center}
            \includegraphics[width=0.9\linewidth]{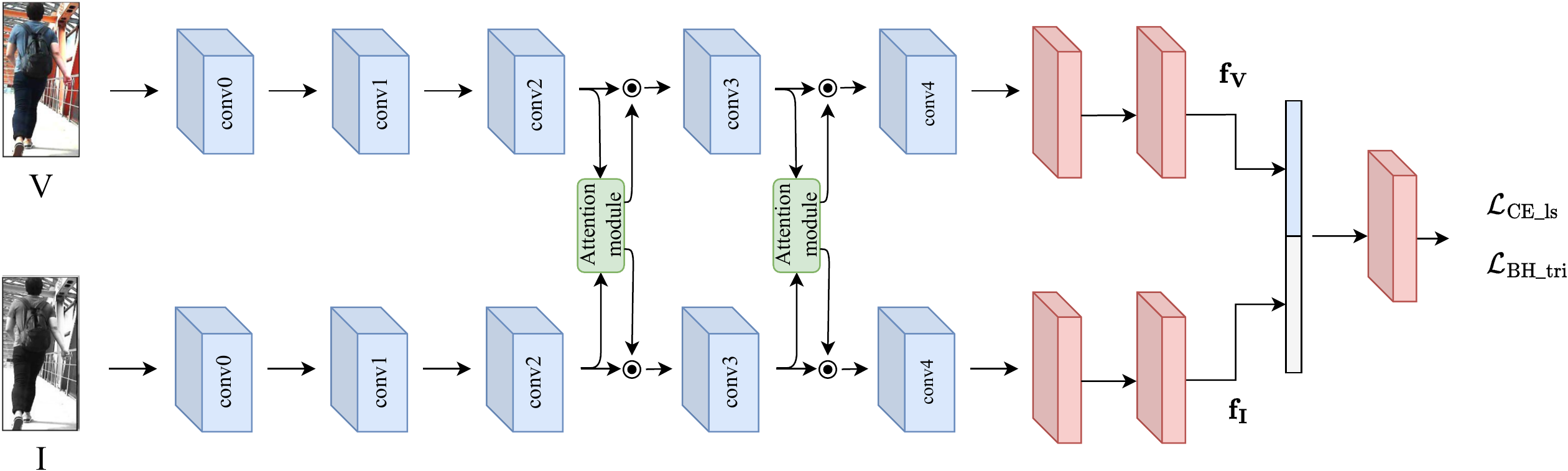}
            \end{center}
            \caption{Learning model architecture for the MMTM and the MSAF approaches while concatenating the feature vectors for fusion. The attention module may be either the MMTM or the MSAF modules.}
            \label{fig:MMTM_MSAF_train}
        \end{figure*}
        
        Modality Attention Network (MAN) \citep{gu2018hybrid} is an attention-based multimodal approach that dynamically weights feature vectors from each modality before fusing them. This model seems meaningful to explore as the dynamic weighting of each modality feature vector should help handle corrupted data. Since the model architecture has been adapted for our person ReID study, its architecture is presented in Fig. \ref{fig:MAN}. 
        
        Two backbones first extract each visible $\mathbf{f_V}\in  \mathbb{R}^{d}$ and infrared $\mathbf{f_I}\in  \mathbb{R}^{d}$ modality features, with $\textit{d}\in \mathbb{R}$. The obtained vectors are concatenated and passed through a modality attention module, which learns to generate soft attention weights. The soft weights allow the model to give more importance to the discriminant modality features in the final embedding. To do so, the concatenation of the two embeddings goes through two dense layers and a final softmax $\sigma$ regression, which produces the soft weights $\text{S}_\text{V} \in \mathbb{R}$ for the visible and $\text{S}_\text{I} \in \mathbb{R}$ for the infrared modalities. Soft weights are produced as follows:  
        \begin{equation}\label{eq:AttentionWeights}
            [\text{S}_\text{V},\text{S}_\text{I}]=\sigma(\mathbf{W_2}\text{tanh}(\mathbf{W_1}[\mathbf{f_V},\mathbf{f_I}]^T + \mathbf{b_1}) + \mathbf{b_2})
        \end{equation}
        where $\mathbf{W_1} \in \mathbb{R}^{k \times d}$ and $\mathbf{W_2} \in \mathbb{R}^{1 \times k}$ are weight matrix, $k\in \mathbb{R}$ being an hyper-parameter, $\mathbf{b_1} \in \mathbb{R}^{k \times 2}$, and $\mathbf{b_2} \in \mathbb{R}^{1 \times 2}$ are biases. 
        
        Thanks to soft attention weights, visible and infrared original features are then weighted, respectively noted $\mathbf{f_V}^w$ and $\mathbf{f_I}^w$. For the visible modality, $\mathbf{f_V}^w=\text{S}_\text{V}\times \mathbf{f_V}$, and for the infrared modality, $\mathbf{f_I}^w=\text{S}_\text{I}\times \mathbf{f_I}$. Then, the predicted output vector $\mathbf{\hat{y}}$ is computed by passing the concatenation or the element-wise sum of the $\mathbf{f_V}^w$ and $\mathbf{f_I}^w$ vectors through a final softmax layer for classification. 
        
        As a consequence of the CL and NCL camera scenarios and the induced spatial alignment, which might influence the feature vector's composition, we also consider the element-wise sum fusion of the feature vectors in this work. Concatenation conserves each feature definition while fusing, but doubles the feature vector dimension. Summation makes the fused vector of the original feature vector size but may erase knowledge if the embedded concepts are not aligned.

        \subsubsection{Multimodal transfer module}\label{sec:MMTM}
        
        The Multimodal Transfer Module (MMTM) \citep{mmtm2020joze} is an approach that focuses on channel attention to refactor the feature maps from two or more modality CNN streams regarding the spatial statistics of each. As the refactoring is done dynamically and based on the statistics of each given input, such attention should also be helpful while facing corrupted data. Two similar backbones are used to extract the features from each V and I representation. Two modules are used for our architecture (Fig. \ref{fig:MMTM_MSAF_train}), after the third and the fourth convolution blocks, allowing for intermediate and high-level feature refactoring. For a given layer $l \in \mathbb{N}$, the visible and the infrared modality feature maps are respectively noted $\mathbf{F_V^\textit{l}} \in \mathbb{R}^{H\times W \times C}$ and $\mathbf{F_I^\textit{l}}\in \mathbb{R}^{ H \times W \times C}$, with $\textit{H} \in \mathbb{R}$, $\textit{W}\in \mathbb{R}$ and $\textit{C}\in \mathbb{R}$ being respectively the feature maps height, width and channel size. The feature map from each stream is first squeezed with a global average pooling layer over the spatial dimension, leading to two linear vectors of channel descriptors. Those vectors are concatenated and passed through a dense layer, following equation \eqref{eq:MMTM_sharedRPZ}, to obtain the joint representation $\textbf{J}^l \in \mathbb{R}^{C_J}$.
        \begin{equation}\label{eq:MMTM_sharedRPZ}
        \mathbf{J}^l = 
        \mathbf{W}([AvgPool(\mathbf{F_V^\textit{l}});AvgPool(\mathbf{F_I^\textit{l}})]) + \mathbf{b}
        \end{equation}
        where $\mathbf{W} \in \mathbb{R}^{C_J \times C^2}$ is a weight matrix, $\mathbf{b} \in \mathbb{R}^{C_J}$ the bias of the dense layer, and $C_J = C^2/4$ to limit the model capacity and increase the generalization power \citep{mmtm2020joze}. 
        Then, an excitation signal is produced with a distinct dense and softmax activation layer applied for each modality to the shared channel descriptor $\mathbf{J}^l$. Finally, this excitation signal is broadcasted through the spatial dimension for each modality with an element-wise product, following equations \eqref{eq:MMTM_excitationAndMultiplication}, forming the final weighted feature maps $\mathbf{F_V^\textit{l}}^w \in \mathbb{R}^{ H \times W \times C}$  and $\mathbf{F_I^\textit{l}}^w \in \mathbb{R}^{ H \times W \times C}$. 
        %
        \begin{equation}\label{eq:MMTM_excitationAndMultiplication}
        \begin{aligned}
        {\mathbf{F_V^\textit{l}}}^w = 2 \times \sigma(\mathbf{W_V}\textbf{J}^l + \mathbf{b_V}) \odot \mathbf{F_V^\textit{l}} \\
        {\mathbf{F_I^{l}}}^w = 2 \times \sigma(\mathbf{W_I}\textbf{J}^l + \mathbf{b_I}) \odot \mathbf{F_I^\textit{l}}
        \end{aligned}
        \end{equation}
        where $\mathbf{W_V} \in \mathbb{R}^{C \times C_J}$ and $\mathbf{W_I} \in \mathbb{R}^{C' \times C_J}$ are weight matrix and $\mathbf{b_V} \in \mathbb{R}^{C}$, $\mathbf{b_I} \in \mathbb{R}^{C}$ the bias of the dense layers. $\sigma$ stands for the sigmoid function. The element-wise product is represented by $\odot$. 
        

 \subsubsection{Multimodal split attention fusion}\label{sec:MSAF}
        
        The Multimodal Split Attention Fusion module (MSAF) proposed by \cite{msaf2020su} also works from the channel attention principle. Modules are applied at the same locations for this model (Fig. \ref{fig:MMTM_MSAF_train}). Let us describe the MSAF module. First, the visible and infrared feature maps $\mathbf{F_V^\textit{l}} \in \mathbb{R}^{H \times W \times C}$ and $\mathbf{F_I^\textit{l}} \in \mathbb{R}^{H \times W \times C}$ are split into $n \in \mathbb{R}$ visible and infrared sub feature maps, respectively noted $\mathbf{S^\textit{l}_V}\in \mathbb{R}^{H \times W \times \frac{C}{n}}$ and $\mathbf{S_I^\textit{l}}\in \mathbb{R}^{H \times W \times \frac{C}{n}}$. The $n$ splits from each modality are element-wise summed and fed to a global average pooling layer to get a global channel descriptor per modality noted $\mathbf{J_V^\textit{l}} \in \mathbb{R}^\frac{C}{n}$ and $\mathbf{J_I^\textit{l}}\in \mathbb{R}^\frac{C}{n}$. Then, the global channel descriptor from each modality is element-wise summed and passed through a dense layer, followed by a batch normalization and a ReLU activation to catch the inter-channel correlations, forming the common channel descriptor $\mathbf{J}^l \in \mathbb{R}^\frac{C}{n}$. From $\mathbf{J}^l$, $n$ excitation signals are produced per modality, using a dense layer and a softmax activation on $\mathbf{J}^l$ for each original feature map split. These excitation signals are then broadcasted through the spatial dimension for each split with an element-wise product, following equations \ref{eq:MSAF_excitationAndMultiplication}, forming the final weighted splits ${\mathbf{S^\textit{l}_{\mathbf{V}_\textit{i}}}}^w \in \mathbb{R}^{H \times W \times \frac{C}{n}}$ and ${\mathbf{S^\textit{l}_{\mathbf{I}_\textit{i}}}}^w \in \mathbb{R}^{H \times W \times \frac{C}{n}}$.
        \begin{equation}\label{eq:MSAF_excitationAndMultiplication}
        \begin{aligned}
            {\mathbf{S^\textit{l}_{\mathbf{V}_\textit{i}}}}^w = \sigma(\mathbf{W_{V_\textit{i}}}\mathbf{J_V^\textit{l}} + \mathbf{b_{V_\textit{i}}}) \odot \mathbf{ \mathbf{S}^l_{V_\textit{i}} } \\
            {\mathbf{S^\textit{l}_{\mathbf{I}_\textit{i}}}}^w = \sigma(\mathbf{W_{T_\textit{i}}}\mathbf{J_I^\textit{l}} + \mathbf{b_{T_\textit{i}}}) \odot \mathbf{S^l_{T_\textit{i}}}
        \end{aligned}
        \end{equation}

        The $n$ excited splits are concatenated together for each modality to get the final weighted feature maps ${\mathbf{F^\textit{l}_{\mathbf{V}_\textit{i}}}^w \in \mathbb{R}^{H \times W \times C}}$ and  $\mathbf{F^\textit{l}_{\mathbf{I}_\textit{i}}}^w \in \mathbb{R}^{H \times W \times C}$. One can notice that the model needs fewer parameters than the MMTM approach, thanks to the feature map splits.

\section{Corrupted Datasets}\label{sec:UCD_CCD}

    \begin{figure*}[htp]
        \centering
        \includegraphics[width=\textwidth]{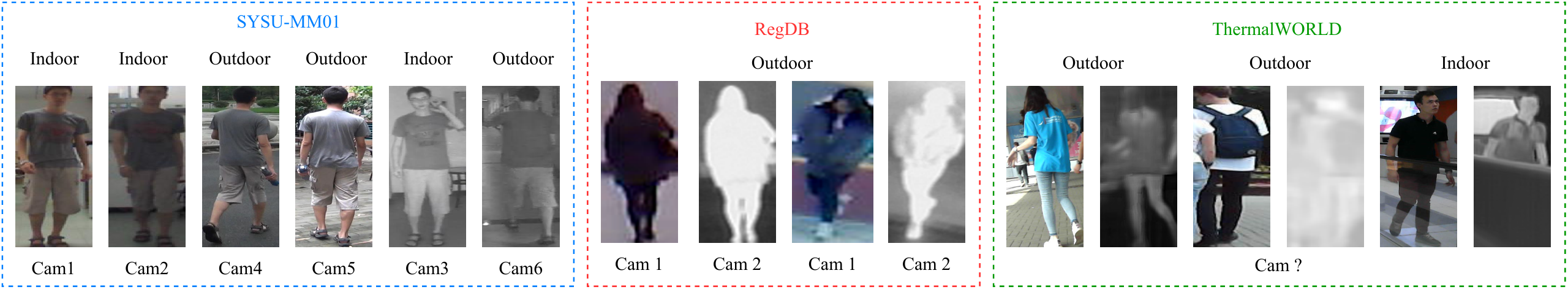}
        \caption{Examples from SYSU-MM01, RegDB and ThermalWORLD. ThermalWorld does not provide camera information.}
        \label{fig:dataset_samples}
    \end{figure*}
    
To better simulate real-world conditions while evaluating a model, the focus has been on corrupted test sets over the last few years \citep{hendrycks2019benchmarking, chen2021benchmarksDataset_C, michaelis2019benchmarking}. However, those benchmark test sets were proposed for single modality settings, whereas our objective is to evaluate the value of V-I multimodal models. As both the V and the I modalities encode from visual cues, corruptions that affect the visual modality may also affect the infrared modality, such as occlusions or weather-related corruptions. From this observation, the 20 visible corruptions from \cite{chen2021benchmarksDataset_C} are extended to the infrared domain in this work, allowing us to provide two corrupted datasets. Those two datasets are suited to the co-located (CL) and the not co-located (NCL) settings. 

In the following sections, the three clean datasets are first detailed. A presentation of the used modality corruptions follows. Finally, our two corrupted datasets are detailed. 

    \begin{table}[htbp]
      \centering
      \fontsize{6}{7.2}\selectfont
      \caption{Datasets statistics. \textbf{V} = \textbf{V}isible and \textbf{I} = \textbf{I}nfrared. Image size and number of samples per identity are presented as: Min;Max;Avg. BRISQUE \citep{mittal2011blindBRISQUE} measure is shown as: avg±std.}
        \begin{tabular}{l||c|c|c}
        \midrule
        \textbf{Statistic} & \textbf{SYSU} & \textbf{RegDB} & \textbf{TWORLD} \\
        \midrule
        V-images  & 29 033 & 4120  & 8125 \\
        I-images  & 15 712 & 4120  & 8125 \\
        V-Camera  & 4     & 1     & 16 \\
        I-Camera  & 2     & 1     & 16 \\
        Cameras setting & NCL & CL   & CL \\
        Identities  & 491   & 412   & 409 \\
        V-images/id & 10;144;59.1 & 10;10;10& 1;155;19.9 \\
        I-images/id & 10;144;32.0 & 10;10;10 & 1;155;19.9 \\
        Image width & 26;1198;111 & 64;64;64 & 10;810;141  \\
        Image height & 65;879;291 & 128;128;128 & 25;897;353 \\   
        V-BRISQUE & 30.50±12.26 & 38.84±9.86 & 27.79±13.28 \\
        I-BRISQUE & 40.52±8.42 & 38.81±9.56 & 60.25±8.67 \\
        \midrule
        \end{tabular}%
      \label{tab:datasets_details}%
    \end{table}%
    
    \subsection{Clean datasets}\label{sec:clean_datasets}
    
    The three used datasets present distinct statistics (Tab. \ref{tab:datasets_details}) suited to build and draw a strong study.
    
    \noindent \textbf{SYSU-MM01} \citep{SYSU} gather 4 visible and 2 infrared cameras, with 491 distinct individuals, $29 033$ RGB, and $15 712$ IR images. The V and I cameras are not co-located, so the scene's spatial description varies from one modality to another for a given V-I image pair. 
    
    \noindent \textbf{RegDB} \citep{RegDB} is a much smaller dataset, with one camera only per modality, the V and I cameras being co-located. A single 10 images tracklet is available per identity and camera. Hence, RegDB $412$ identities lead to $4 120$ images per modality.  
    
    \noindent \textbf{ThermalWorld\footnote{Download \href{link}{https://drive.google.com/file/d/1XIc_i3mp4xFlDJ_S5WJYMJAHq107irPI/view} obtained from github ThermalGAN \href{issues}{https://github.com/vlkniaz/ThermalGAN/issues/12}.}} \citep{ThermalWORLD} has only its training part available, leading us to $409$ distinct identities. 16 co-located cameras per modality captured each $8125$ image. However, the infrared images are of terrible quality, with a BRISQUE \citep{mittal2011blindBRISQUE} value of $60.25$, much higher than RegDB and SYSU-MM01 ones, being at $38.81$ and $40.52$ respectively.  


\begin{figure*}[htp]

    \centering
    \includegraphics[width=.3\textwidth]{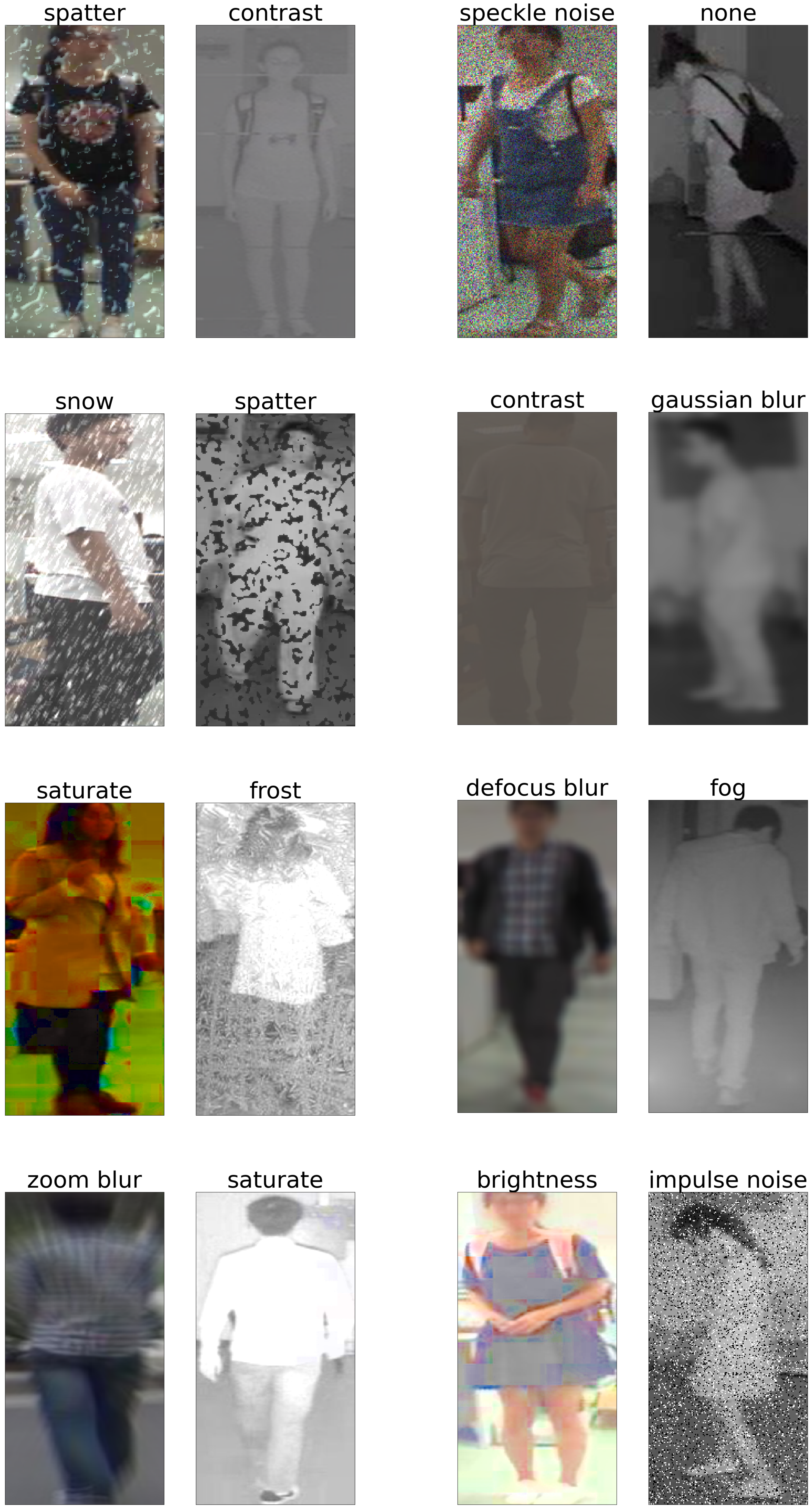}
    \hfill
    \includegraphics[width=.3\textwidth]{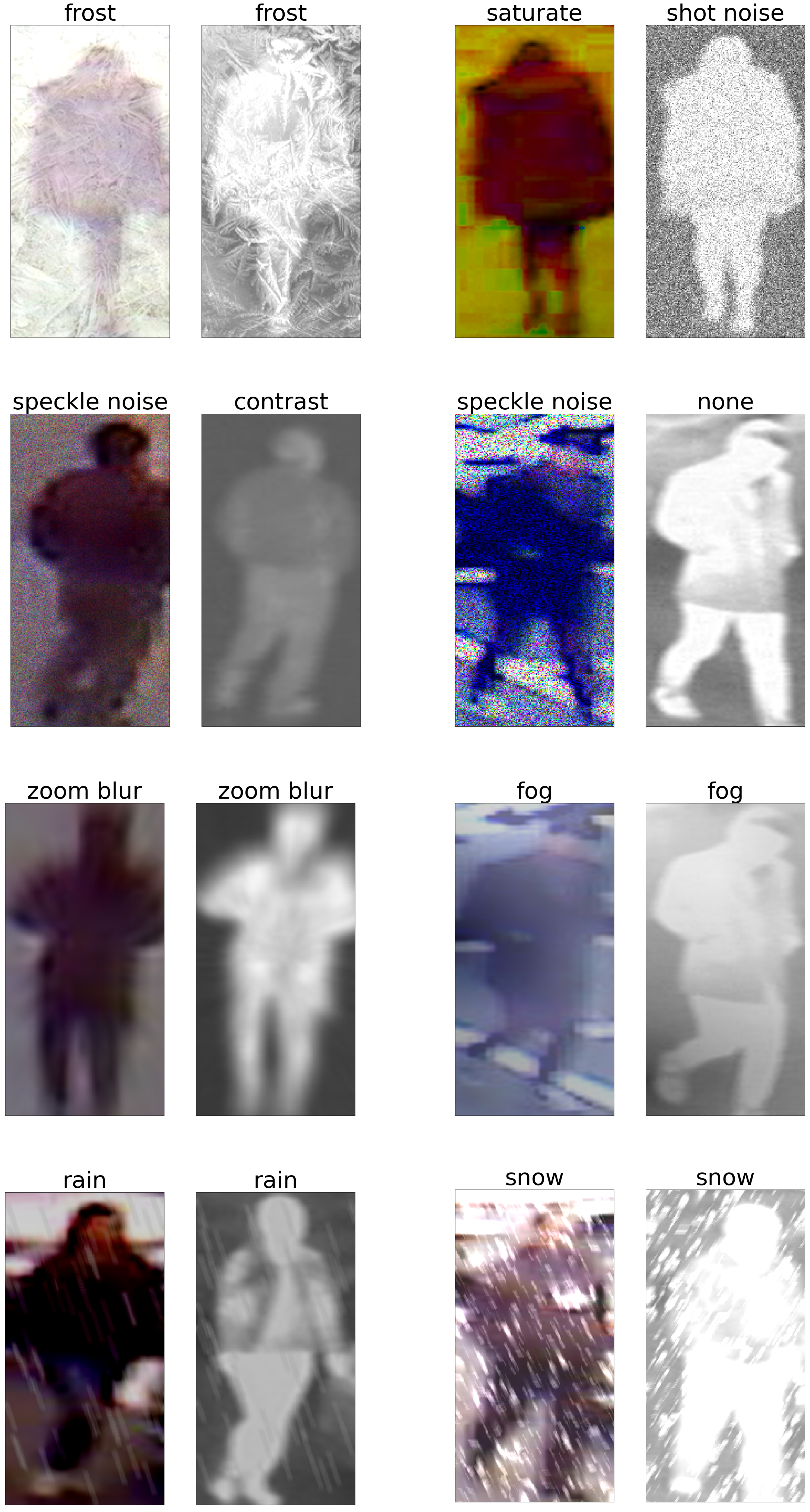}\hfill
    \includegraphics[width=.3\textwidth]{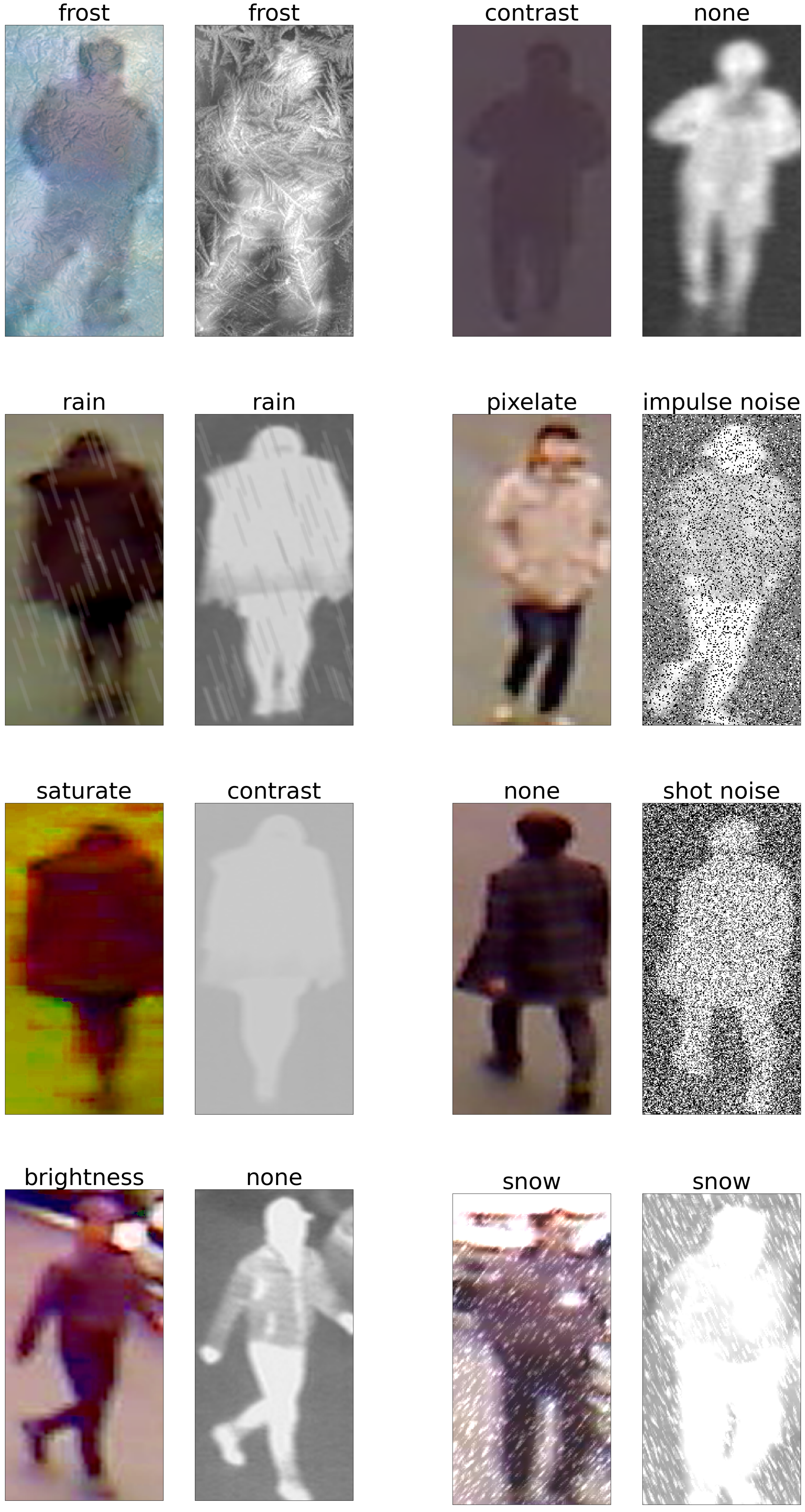}
    \subfloat[\label{subfig:UCD}SYSU-MM01-UCD]
    {\hspace{.3\linewidth}}\hfill
    \subfloat[\label{subfig:CCD}RegDB-CCD]
    {\hspace{.3\linewidth}}\hfill
    \subfloat[\label{subfig:CCD-50}RegDB-CCD-50]{\hspace{.3\linewidth}}
    \caption{Samples from our three corrupted datasets. Visuals do not represent all available dataset versions, as each dataset has its own UCD, CCD, and CCD-50 version.}
    \label{fig:corrupted_dataset_samples}

\end{figure*}
    
    \subsection{Modality corruptions}\label{app:thermal_modality_corruptions}
    \cite{hendrycks2019benchmarking,chen2021benchmarksDataset_C} used 20 corruptions of the visual modality, which were regrouped into four distinct types - noise, weather, blur, and digital. In this work, the used corruptions are the same for the visual modality. However, the I modality can also be affected by multiple corruptions, which is considered. In fact, 19 of the corruptions affecting the visual modality can also apply to the infrared with a few slight adjustments (Corruptions taxonomy figure and corruptions adjustments table in the appendix 1.A). 
    
    First, the current luminosity does not impact the IR modality, so brightness corruption is not used for this modality. Then, different noises, like Gaussian, Shot, Impulse, and Speckle, are applied similarly, except each noise is turned into grayscale values to respect the infrared modality single color channel encoding. Spatter and frost are two other corruptions that needed to be grayscaled before being applied to the infrared images. Indeed, blue-colored water or brown-colored dirt was applied for spatter, and frozen blue masks for frost. As a last adjustment, the saturation is expressed differently for the I modality, visually brightening the object of interest eventually if this one is too close to the camera, instead of modeling color intensity for the visual modality. Finally, all other corruptions were applied similarly for the V and I modalities.


\subsection{Uncorrelated corruption dataset}\label{sec:UCD_description}

    The Uncorrelated Corruption Dataset (UCD) is proposed as a first way to evaluate the models' corruption robustness. To build UCD, the corruptions are randomly and independently selected and applied on each modality for a given V-I test pair, making it highly challenging. The camera corruption independence from V to I modality is suited for a NCL camera setting, as it is the case for SYSU-MM01. Indeed, for example, a visible indoor and an outdoor infrared camera would lead to weather appearing only on the infrared camera or to blur, impacting one camera only while the other is impacted independently. As applied corruptions are most of the time distinct from one modality to another under UCD, it should allow each modality to compensate for the corrupted features from the other. Hence, this setting should be a great way to evaluate the models' ability to select the information of interest from one or another modality. !
    
    \subsection{Correlated corruption dataset}\label{sec:CCD_description}
    
    One can expect some corruption to be correlated from one camera to another, corruption type-wise as intensity-wise. As a brief example, the rain is expected to appear on both visible and infrared cameras simultaneously, especially if those are co-located. However, some other types of corruption, such as image saturation, are camera dependent and would happen punctually on one camera with no correlation with the other. The CCD dataset is proposed from these observations, suited for CL cameras and gathering the following characteristics (Tab. \ref{tab:correlated_dataset_precisions}). 
    
    At first, weather-related corruptions such as fog, rain, frost, and snow appear much correlated, so the weather from one camera is assumed to appear with the same level of corruption on the other. Spatter expresses the water or dirt splashes on the cameras, which has a great chance to happen on both cameras considering co-located cameras, but with a level that might differ; the level is selected randomly and independently. Similar behaviors for blur-related corruptions would also make sense in real-world conditions if cameras are co-located since those corruptions are a consequence of camera settings, like exposure time or focus, for example, but which also mostly depends on the current scene. Because each modality camera might be more or less reactive regarding the situation, we consider that blur-related corruptions (i.e., defocus, gaussian, glass, zoom, motion blurs)  affect the two modalities simultaneously with an intensity level that can differ. The intensity level is randomly and independently selected except for motion blur corruption. Indeed, infrared cameras usually have a higher exposure time than visible cameras, making those more affected by motion blur. Consequently, the level is always selected as equal or superior for the infrared modality compared to the visible one. 

    \begin{table}[htbp]
      \centering
      \caption{Correlated (center) and uncorrelated (right) corruptions  are presented, along with the relation between levels of corruption (left) from the V to the I modality for correlated corruptions. }
        \begin{tabular}{c|l||l}
        \midrule
        \multicolumn{1}{c|}{\textbf{Level}}
        & \multicolumn{1}{c||}{\textbf{Correlated}} & \multicolumn{1}{c}{\textbf{Uncorrelated}} \\
        \midrule

        V = I   &  Fog & Gaussian noise \\
        V = I & Frost & Shot noise \\
        V = I  & Snow & Impulse noise \\
        V = I  & Rain & Speckle noise  \\
        V $\neq$ I  & Spatter & Elastic transform \\
        V $\neq$ I & Defocus blur & Saturation  \\
        V $\neq$ I & Gaussian Blur & JPEG compression \\
        V $\neq$ I & Glass Blur & Pixelate  \\
        V $\neq$ I & Zoom Blur & Contrast \\
        V $\leq$ I & Motion Blur & Brightness \\

        \midrule
        \end{tabular}%
      \label{tab:correlated_dataset_precisions}%
    \end{table}%
    
	Concerning the ten remaining corruptions, those are much related to data encoding and can affect visible or infrared cameras independently. The hetero-modality is consequently corrupted if the selected corruption lies in the correlated corruptions. Otherwise, we randomly apply another corruption among the uncorrelated ones to the hetero-modality. Considering modalities as always corrupted is an extreme scenario, which is attractive to frame models' behaviors but not entirely realistic. Hence, the UCD-X dataset is proposed. In this configuration, $X$\% of the corrupted pairs affected by uncorrelated corruptions are formed with one of the two modalities remaining clean. In practice, we fixed it at $50\%$, but this value can be tweaked to make the datasets more or less challenging for further experiments.


\section{Multimodal Data Augmentation}\label{sec:ML_MDA}

    \begin{figure*}[t]
            \begin{center}
            \includegraphics[width=0.9\linewidth]{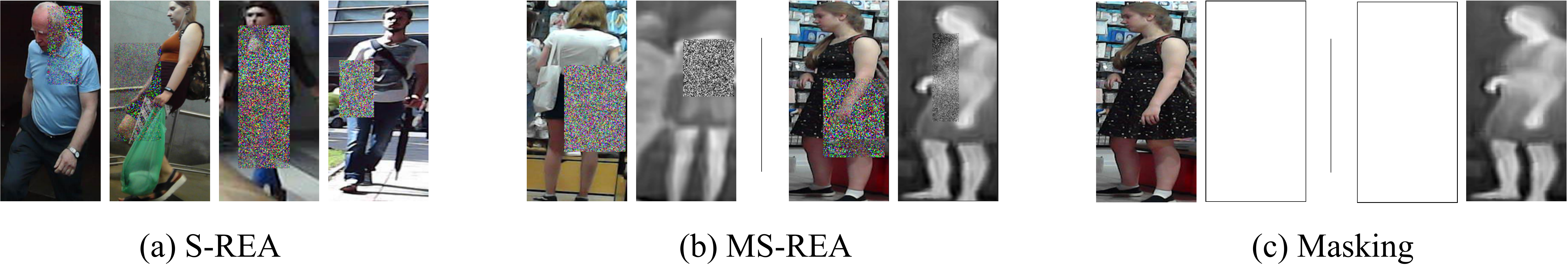}
            \end{center}
            \caption{Soft random erasing (S-REA) \citep{chen2021benchmarksDataset_C} and our MDA based on multimodal soft random erasing (MS-REA) and modality masking.}
            \label{fig:DA}
    \end{figure*}
        
    The explored models are based on co-learning, allowing each modality stream to adapt to the other one \citep{baltruvsaitis2018multimodal, rahate2022multimodal}. We propose a new MDA approach, the Masking and Local Multimodal Data Augmentation (ML-MDA), for better learning of the models. In practice, ML-MDA is based on two components: multimodal soft random erasing (MS-REA) and modality masking (Fig. \ref{fig:DA}). Those two data augmentations are used together during the learning process to make the learned co-learning model robust and accurate in a challenging inference environment. \\ 
    
    \subsection{Multimodal soft random erasing}
    
    Making a multimodal model focus on modality-specific features is challenging, as the model usually mainly focuses on shared features \citep{baltruvsaitis2018multimodal}. Augmenting the multimodal data with local occlusions may help the model to emphasize modality-specific feature importance, as some features will be available only from one or another modality. The soft random erasing \citep{chen2021benchmarksDataset_C} (S-REA) (Fig. \ref{fig:DA}.a.) uses local occlusions to learn the model not to rely only on the most important features, but consider unimodal learning and consequently not exploit this aspect. 
    
    The MS-REA data augmentation is proposed to close this gap. Instead of replacing a proportion of the pixels in a given image patch with random pixel values for the visible modality only as S-REA, MS-REA applies a patch on both the visible and infrared modalities. Grayscaled random pixel values are used for patches on the thermal modality to respect the infrared thermal image definition, encoded on one channel, and potentially aligning better with real-world corruptions. The spatial patch location is randomly and independently selected from the visible and infrared images for a given V-I pair. To close the occlusion gap brought by the applied patches through MS-REA, the model must learn how to select each modality feature when partial information is available from each modality. Such behavior is expected to extend well to real-world corruption. 
    
    \subsection{Modality masking}
    
    A modality might be punctually unavailable or primarily uninformative. Thus, the model shall learn how to cancel a modality to reduce its impact on the final prediction. The modality masking approach is expected to allow it by punctually replacing one or another modality with an entirely blank image. Instead of masking the multimodal representation as it has been done by \cite{gabeur2022masking}, a representation is extracted from the masked input, so the model has to learn how to cancel its influence on the final results. This also forces the model to focus more on modality-specific features since one modality only contains all the meaningful knowledge for ReID. This DA should supposedly complement the previously presented MS-REA approach by balancing each modality's importance in the final embedding regarding each modality level of corruption, whereas MS-REA should learn the model to select the features within each modality better. MS-REA should also make models' put more emphasis on the modality-specific features, this time thanks to the independent occlusions locations on each image.

\section{Results and Discussion}\label{sec:results}

\subsection{Experimental methodology}\label{sec:implementationDetails}
 
    \noindent \textbf{Data division.} SYSU-MM01 and RegDB datasets have well-established V-I cross-modal protocols \citep{Join_Pixel_align, wang2019learning, ye2019bi}, but multimodal protocols were not existing prior to our preliminary work \citep{our_preliminary}. Following them again, 395 and 96 identities from SYSU-MM01 are respectively used for the training and the testing set. For RegDB, 412 identities are divided into two identical sets of 206 individuals for learning and testing. The SYSU-MM01 train/test ratio is kept for ThermalWORLD, leading to 325 training identities and 84 for testing. A 5-fold validation \citep{K-cross-validation} is performed over the data used for training, using folds of respectively $79$, $41$, and $65$ distinct identities for SYSU-MM01, RegDB, and ThermalWORLD.
    
    \noindent \textbf{Data augmentation.} Our proposed multimodal extensions MS-REA is used with the same appearance augmentation probability as S-REA \citep{chen2021benchmarksDataset_C}. Modality Masking is applied randomly on one or another modality, with equiprobability, and occurs with a default probability of $1/8$. When used on unimodal models, the CIL \citep{chen2021benchmarksDataset_C} DA is used the same way as the original authors. For the RegDB dataset only, the validation set is given the same DA as the training set as the maximum performances were reached in the early epochs otherwise. This way, better convergence was observed, allowing learning complex cues by the model.
    
    \noindent \textbf{Pre-processing.} A data normalization is done at first by re-scaling RBG and IR images to $144 \times 288$. Random cropping with zero padding and horizontal flips are adopted for base DA. Those parameters were proposed by \cite{PersonREID_outlook} on RegDB and SYSU-MM01 datasets. The same normalization is kept under ThermalWORLD for consistency among protocols.  
    
    \noindent \textbf{Performance measures.} The mean Average Precision (mAP), and the mean Inverse Penalty (mINP) are used as performance measures, commonly used for person ReID \citep{PersonREID_outlook}. The mAP is the mean computed over all query image ratio of retrieved matches over total matches. However, mAP does not reflect the worst-case scenario, unlike the mINP measure, which applies a penalty on the hardest matches, making it a great complementary measure.
    
    \noindent \textbf{Hyperparameters.} The hyperparameters values in our models were set based on the default AGW \citep{PersonREID_outlook} baseline. The SGD is used for training optimization, combined with a Nesterov momentum of $0.9$ and a weight decay of $5e-4$. Our models are trained through $100$ epochs. Early stopping is applied based on validation mAP performances. The learning rate is initialized at $0.1$ and follows a warming-up strategy \citep{warming_up_strat}. The batch size is 32, with 8 distinct individuals and 4 images per individual. The paired image is selected by default for RegDB and ThermalWORLD. For the SYSU-MM01 dataset, the images from the hetero modality are randomly selected through the available ones to form a pair for a given identity. 
    
    \noindent \textbf{Losses.} The Batch Hard triplet loss \citep{TripletLoss} $\mathcal{L}_{\text{BH\_tri}}$ and the cross-entropy with regularization via Label smoothing \citep{szegedy2016rethinking} $\mathcal{L}_{\text{CE\_ls}}$ are used as loss functions for our models. Indeed, the former is widely used in person ReID approaches \citep{Join_Pixel_align,Hi-cmd,PersonREID_outlook}, so the same margin value is fixed at $0.3$, and the latter is part of the CIL implementation \citep{chen2021benchmarksDataset_C}. The total loss corresponds to the sum of both losses. The batch hard triplet loss aims at reducing the distance in the embedding space for the hardest positives while increasing the distance for the hardest negatives. The regularization with label smoothing reduces the gap between logits, making the model less confident in predictions and hence improving generalization \citep{muller2019does}.

    \noindent \textbf{Models details.} 
    MMSF is used with $\textit{l}=4$ for NCL and $\textit{l}=0$ for CL cameras (Appendix 1.B). The influence of concatenation or sum of the feature vectors is explored in the Appendix 1.C and allowed to converge to use MMTM S (Sum) and MSAF C (Concatenation) for RegDB, and MMTM C and MSAF S for ThermalWORLD and SYSU-MM01.
     
    \noindent \textbf{Leave-one-out query strategy.} \label{LOO}
    The Leave-One-Out Query (LOOQ) strategy, proposed in our preliminary work \citep{our_preliminary}, is used the same way in this study. The LOOQ treats the extreme but meaningful case in which one would have only a unique image of the person to ReID and multiple footages containing images of this same person in the gallery. Every pair of images is alternatively used as a probe set while all the other pairs join the gallery. While an interesting evaluation strategy, this also allows us to respect the original dataset statistics (Tab. \ref{tab:datasets_details}) by authorizing the number of used gallery images per individual to vary.   
\subsection{Scenario with not co-located cameras} \label{sec:benchmark_multimodal_NCL}

   Not co-located (NCL) V-I cameras imply that a pair of images for a given individual is built from two distinct viewpoints. Consequently, images in a given V-I pair will not be spatially aligned from one modality to another. Having two viewpoints for a given V-I pair should allow more cues and be more discriminant to ReID than a co-located (CL) setting. Indeed, if the person is occluded or partially visible from one camera modality, for example, the hetero-modality camera might have a better view and compensate for the missing features. However, correlations from one modality to another may be harder to find for NCL cameras as the scene appears much different between modalities \citep{wang2021role}. For example, the spatial information remaining in the features when the fusion is done may act as noise for the model due to the absence of alignment. 
   
   Since various corruptions can impact either modality, a multimodal model might be disturbed by the supplementary modality and could consequently be less able to ReID than a well-trained single-modal model. The upcoming study is proposed to determine whether or not the multimodal framework is worthwhile given the above statements and to seek the best approach to follow.
   
   


\subsubsection{Robustness to corruption}\label{sec:Models_corruption_robustness_NC}

        \begin{table*}[htbp]
      \centering
      \caption{Unimodal and multimodal models performances while evaluated on clean and corrupted SYSU-MM01 datasets. Unimodal V and I stands respectively for unimodal visible and thermal models. In bold and blue are the first and second best approaches respectively.}
        \begin{tabular}{cl||cc|cc|cc|cc}
        \midrule
        
              & \textbf{Model} & \multicolumn{2}{c|}{\textbf{Clean}} & \multicolumn{2}{c|}{\textbf{UCD}} & \multicolumn{2}{c|}{\textbf{CCD}} & \multicolumn{2}{c}{\textbf{CCD-50}} \\
              &       & \textbf{mAP} & \textbf{mINP} & \textbf{mAP} & \textbf{mINP} & \textbf{mAP} & \textbf{mINP} & \textbf{mAP} & \textbf{mINP} \\
        \midrule
        \multirow{8}[2]{*}{\begin{sideways}No DA\end{sideways}} & Unimodal V & 86.72 & 41.16 & \textbf{32.16} & 1.86 & \textbf{32.11} & \textcolor[rgb]{ .188,  .329,  .588}{1.89} & \textbf{37.70} & 2.15 \\
              & Unimodal I & 77.06 & 30.44 & 13.97 & 1.25  & 13.51 & 1.25  & 18.26 & 1.31 \\
              & Baseline S   & 95.96 & 71.14 & 22.55 & 1.82  & 19.26 & 1.71  & 31.90 & \textcolor[rgb]{ .188,  .329,  .588}{2.37} \\
              & Baseline C & 96,47 & 73,69 & 25.01 & \textcolor[rgb]{ .188,  .329,  .588}{1.90}  & 24.35 & 1.86  & 31.24 & 2.26 \\
              & MAN   & 91.05 & 55.00 & \textcolor[rgb]{ .188,  .329,  .588}{27.76} & 1.84 & \textcolor[rgb]{ .188,  .329,  .588}{27.72} & 1.84 & 33.99 & 2.17 \\
              & MMTM  & 95.71 & 71.35 & 20.00 & 1.59  & 18.31 & 1.70  & 30.59 & 2.25 \\
              & MSAF  & \textcolor[rgb]{ .188,  .329,  .588}{96.77} & \textcolor[rgb]{ .188,  .329,  .588}{77.27} & 25.64 & \textbf{2.03}  & 21.77 & \textbf{1.93}  & \textcolor[rgb]{ .188,  .329,  .588}{34.58} & \textbf{2.54} \\
              & MMSF  & \textbf{97.80} & \textbf{80.93} & 22.23 & 1.65  & 17.70 & 1.60  & 31.15 & 2.12 \\
        \midrule
        \multicolumn{1}{c}{\multirow{8}[1]{*}{\begin{sideways}ML-MDA / CIL\end{sideways}}} & Unimodal V & 86.72 & 42.70 & 52.37 & 3.89  & 52.58 & 3.93  & 55.48 & 4.67 \\
              & Unimodal I & 78.33 & 35.41 & 33.38 & 2.32  & 32.78 & 2.32  & 36.26 & 2.39 \\
              & Baseline S   & 96.54 & 74.49 & 64.00 & 9.72  & 61.81 & 7.53  & 64.69 & 8.26 \\
              & Baseline C & 96.77 & 76.01 & 63.40 & 9.51  & 61.94 & 7.72  & \textcolor[rgb]{ .188,  .329,  .588}{65.79} & 8.71 \\
              & MAN   & \textcolor[rgb]{ .188,  .329,  .588}{97.13} & \textcolor[rgb]{ .188,  .329,  .588}{77.91} & 63.50 & 8.24  & 61.87 & 6.39  & 64.75 & 7.10 \\
              & MMTM  & 95.81 & 74.23 & 64.41 & \textbf{11.49} & 62.30 & \textbf{8.55}  & 64.91 & \textbf{9.22} \\
              & MSAF  & 96.36 & 73.70 & \textbf{67.78} & 10.09 & \textbf{65.49} & \textcolor[rgb]{ .188,  .329,  .588}{8.00} & \textbf{68.91} & \textcolor[rgb]{ .188,  .329,  .588}{9.14} \\
              & MMSF  & \textbf{97.66} & \textbf{79.52} & \textcolor[rgb]{ .188,  .329,  .588}{65.24} & \textcolor[rgb]{ .188,  .329,  .588}{10.41} & \textcolor[rgb]{ .188,  .329,  .588}{63.16} & {7.44} & 65.58 & {8.36} \\
        \midrule
            
        \end{tabular}%
      \label{tab:unimodal_to_mmodal_perf_analysis_NC}%
    \end{table*}%
  
    Multimodal models are compared while evaluated on each clean, UCD, CCD, and CCD-50 version of the SYSU-MM01 evaluation data. Clean data is important as a reference, observing performances under the best-case scenario. UCD and CCD should complete each other. The former will allow observing how the models can adapt and select information from differently corrupted V-I inputs. The latter will present how the models can deal with similarly corrupted inputs, which should make the task harder as it should happen more often that the same features for a given pair get corrupted from V to I. Finally, the CCD-50 should be the easiest evaluation set, with $50$\% of the pairs having one over two modalities remaining clean. This last set should allow observing if some models better deal with punctual unilateral corruption. 

      \noindent \textbf{A) Natural models corruption robustness.} 
    
   To begin with, the models are trained without any data augmentation technique and evaluated on the original and corrupted versions of SYSU-MM01 (Upper half Tab. \ref{tab:unimodal_to_mmodal_perf_analysis_NC}). The considered unimodal models are fine-tuned ResNet-18 models, trained from visible (unimodal V) or infrared (unimodal I) modality only.
    
    Before seeking the models' robustness to corruption, observing good performance on clean evaluation data is essential. In practice, each multimodal approach improves over the unimodal models. From the unimodal to the multimodal setting, the greatest improvement comes between the unimodal visible and the proposed MMSF approach, improving the mAP and mINP percentile point (PP) by $11,08$ and $39,77$, respectively. The impressive performance improvement shows how the infrared modality and the NCL cameras through clean SYSU-MM01 strongly benefit the multimodal ReID.
    
    For each corrupted test set (i.e., UCD, CCD, and CCD-50), as both modalities can impact the ReID either way, one can observe here that each multimodal model (learned without a specific strategy) is less efficient than the unimodal V specialist. Indeed, the unimodal V model reaches $32.16$\% mAP, followed by MAN at $27.76$\% mAP for SYSU-MM01-UCD, for example. 
    
    Focusing on the corrupted datasets, the multimodal models globally reach lower performances from UCD to CCD as expected, with, for example, the MMSF model being respectively at $22.23\%$ or $17.70\%$ mAP. From CCD to CCD-50, one can see that some models seem to react better to unilateral corruptions, as the mAP improvement in PP for the proposed MMSF is about $13.45$, for MSAF about $12.81$, and for Baseline C about $6.89$. 
    
    Among multimodal models, the ranking is inconsistent, from the clean to the corrupted setting. On clean data, the proposed MMSF model presents the highest performances, with mAP about $97.80$\% and mINP about $80.93$\%, closely followed by the attention-based MSAF approach, reaching $96.77$\% mAP and $77.27$\% mINP. On corrupted data, MAN and MSAF appear as better at handling corruption than MMSF. 
    
    From there, it would be hard to advise one or another model with the aim of ReID under real-world conditions. Indeed, the evaluated multimodal models are shown not to learn how to select the right modality information in the face of corruption without using a corruption-dedicated learning strategy. \\

    \noindent \textbf{B) DA impact on models robustness.} 
    
    Performances for the unimodal specialists learned using CIL and the multimodal models learned using ML-MDA are presented in the bottom half of Tab. \ref{tab:unimodal_to_mmodal_perf_analysis_NC}. 
   
   The CIL use for unimodal models increases the models' performances on clean data. Indeed, especially for the unimodal infrared model, mAP and mINP are respectively improved by $1.27$ PP and $4.97$ PP for example. ML-MDA has an impact that is model dependant. Still, most models conserve similar mAPs, except for the baseline sum and MAN that see it considerably improve.
   

    Considering corrupted evaluation sets, using DA brings an impressive corruption robustness improvement to every model. The unimodal V model under UCD improves, for example, from $32.16\%$ to $52.37\%$ mAP, or the baseline C model from $24.35$\% to $61.94$\% mAP under the CCD set. Similar improvements are happening under each corrupted setting.
    
    Using ML-MDA, the multimodal models' performances are now ahead of the unimodal ones by a strong margin on corrupted datasets. Indeed, the models learn to select information from each corrupted modality way better. Also, its usage brings more consistency from the clean to the corrupted setting, making it essential to handle real-world conditions. For example, the proposed MMSF model was, and remains, the most discriminant approach under clean data but is now the second-best approach on corrupted datasets. In contrast, this one came in the fifth position at best without DA.
    
    One may wonder if the multimodal models benefit more from clean data pairs than the unimodal specialists. Results from CCD to CCD-50 (that has 50\% of its pairs containing one clean modality) should help for this analysis. In fact, the mAP gap from unimodal V to MSAF increases by $0.62$ percentage points from CCD to CCD-50, and decreases from unimodal V to MMSF by $0.48$ points. Hence, the multimodal setting seems to benefit globally as much from the clean data pairs as the unimodal V model. However, as 50\% of V-I pair have a clean image, it means 25\% of the data is clean for the unimodal V, which shows, in a way, that the multimodal models are benefiting less from a clean modality but keep up with the unimodal V model thanks to the doubled amount of clean pairs. 
    For deeper analysis, each corruption impact and unilateral corruption are further explored in the next section, so as a qualitative analysis through class activation maps (CAMs) generation (Appendix 1.E.).


     \begin{table*}[htbp]
      \centering
      \fontsize{6}{7.2}\selectfont
  \caption{Corruption-wise performance comparison between unimodal, MSAF, and MMSF models and while corrupting one or the other modality only. Models were trained using DA. In Red are visible model performances without corruption and multimodal models performances that get lower than those due to thermal corruption. In blue are thermal model performances without corruption and models that get lower due to an RGB corruption. JPEG cpr = JPEG compression. Elastic trsf = elastic transform.}
    \begin{tabular}{l||cc|cc|cc|cc|cc|cc}
    \toprule
     & \multicolumn{6}{c|}{\textbf{V Corrupted}}   & \multicolumn{6}{c}{\textbf{I Corrupted}} \\
    \textbf{Corruption} & \multicolumn{2}{c|}{\textbf{Unimodal V}} & \multicolumn{2}{c|}{\textbf{MSAF}} & \multicolumn{2}{c|}{\textbf{MMSF}} & \multicolumn{2}{c|}{\textbf{Unimodal I}} & \multicolumn{2}{c|}{\textbf{MSAF}} & \multicolumn{2}{c}{\textbf{MMSF}} \\
          & \textbf{mAP} & \textbf{mINP} & \textbf{mAP} & \textbf{mINP} & \textbf{mAP} & \textbf{mINP} & \textbf{mAP} & \textbf{mINP} & \textbf{mAP} & \textbf{mINP} & \textbf{mAP} & \textbf{mINP} \\
    \midrule
        No corruption & \textcolor{red}{86.72} & \textcolor{red}{42.70} & 96.36 & 73.70 & 97.66 & 79.52 & \textcolor{blue}{78.33} & \textcolor{blue}{35.41} & 96.36 & 73.70 & 97.66 & 79.52 \\
    Gaussian noise & 73.88 & 23.58 & 93.49 & 62.30 & 95.97 & 70.89 & 43.82 & 6.07  & 90.90 & 50.72 & 92.20 & 55.58 \\
    Shot noise & 79.53 & 30.95 & 94.75 & 67.63 & 96.85 & 75.42 & 43.97 & 6.48  & 91.07 & 51.43 & 92.09 & 55.23 \\
    Impulse noise & 73.46 & 23.44 & 93.22 & 61.46 & 95.91 & 70.54 & 36.78 & 4.26  & 89.87 & 47.89 & 90.44 & 49.53 \\
    Speckle noise  & 82.76 & 35.37 & 95.53 & 70.57 & 97.31 & 77.67 & 53.22 & 10.31 & 92.84 & 57.31 & 93.96 & 62.02 \\
    Defocus blur & 75.03 & 23.68 & 94.75 & 67.08 & 96.58 & 73.83 & 59.68 & 13.56 & 94.21 & 63.25 & 95.58 & 68.22 \\
    Glass blur  & 80.12 & 30.75 & 95.42 & 69.74 & 97.08 & 76.53 & 62.38 & 15.83 & 94.58 & 65.26 & 96.08 & 70.89 \\
    Motion blur  & 79.52 & 29.55 & 95.21 & 68.63 & 97.03 & 76.16 & 65.51 & 17.03 & 94.87 & 65.55 & 96.56 & 72.79 \\
    Zoom blur  & 76.50 & 26.27 & 94.73 & 67.55 & 96.48 & 73.91 & 54.17 & 11.24 & 93.15 & 59.67 & 94.73 & 64.56 \\
    Gaussian blur & 74.51 & 22.79 & 94.66 & 66.55 & 96.50 & 73.21 & 59.43 & 13.33 & 94.09 & 62.87 & 95.40 & 67.80 \\
    Snow  & 36.18 & 4.88  & 85.46 & 43.74 & 85.02 & 41.23 & 5.87  & 1.15  & \textcolor{red}{81.77} & \textcolor{red}{33.67}& \textcolor{red}{46.99} &\textcolor{red}{ 5.16} \\
    Frost & 28.08 & 2.36 & 83.09 & 38.02 & 77.10 & 28.61 & 14.82 & 1.46 & \textcolor{red}{85.30} & \textcolor{red}{38.31} & \textcolor{red}{74.31} & \textcolor{red}{21.06} \\
    Fog   & 34.22 & 3.82 & 84.48 & 40.91 & 87.71 & 42.69 & 16.23 & 1.43 & \textcolor{red}{86.36} & \textcolor{red}{40.06} &\textcolor{red}{ 79.71} & \textcolor{red}{27.57} \\
    Brightness & 66.96 & 16.83 & 92.46 & 59.21 & 94.69 & 64.89 & /     & /     & /     & /     & /     & / \\
    Rain  & 50.66 & 7.99 & 87.86 & 47.65 & 89.85 & 51.28 & 31.79 & 2.85 & 89.56 & 47.24 & \textcolor{red}{86.52} & \textcolor{red}{41.69} \\
    Spatter  & 70.28 & 21.06 & 93.08 & 61.45 & 95.09 & 67.49 & 29.18 & 3.27 & 88.34 & 45.37 & \textcolor{red}{78.86} & \textcolor{red}{32.04} \\
    Contrast & 23.00 & 1.45 & \textcolor{blue}{77.64} &\textcolor{blue}{ 29.44} & \textcolor{blue}{78.08} & \textcolor{blue}{26.19} & 33.04 & 2.90 & \textcolor{red}{85.32} & \textcolor{red}{37.07} & 88.18 & 39.89 \\
    Elastic trsf & 75.67 & 26.93 & 94.50 & 66.67 & 96.39 & 73.27 & 42.88 & 6.43 & 92.16 & 55.63 & 92.46 & 55.91 \\
    Pixelate  & 84.62 & 37.65 & 96.24 & 73.24 & 97.70 & 79.99 & 73.35 & 25.02 & 96.01 & 71.51 & 97.47 & 78.46 \\
    JPEG cpr & 69.43 & 18.43 & 93.69 & 62.20 & 95.32 & 67.99 & 70.08 & 20.83 & 95.47 & 68.53 & 96.92 & 75.24 \\
    Saturation  & 66.39 & 16.59 & \textcolor{blue}{72.06} & \textcolor{blue}{23.99} & \textcolor{blue}{57.25} & \textcolor{blue}{10.71} & 45.61 & 5.83 & 90.85 & 51.01 & 93.58 & 59.07 \\
    \bottomrule
    \end{tabular}%
      \label{tab:individual_corruption_impact}%
    \end{table*}%

    \subsubsection{Specific corruption impact}\label{sec:uni_corrupt}

    Corruption can sometimes be one-sided, as with NCL cameras or digital corruption. Hence, we may wonder whether some corruptions of the infrared modality will make the unimodal V model advantageous against multimodal models. This question is also raised for visible corruptions and the unimodal I specialist against multimodal models. To answer those, performances of the unimodal visible and thermal models and those of MSAF and MMSF are observed regarding each corruption (Tab. \ref{tab:individual_corruption_impact}) while corrupting only either modality. The MSAF and the proposed MMSF models are selected as those that performed the best over the evaluated multimodal models.
    
    Weather-related corruptions are the most challenging over the infrared modality, reflected in the lower MSAF and MMSF performances under those data alterations. Compared to the unimodal V model, MSAF is under for 4 corruptions and MMSF for 5 (in red), and both are, on average, much higher for the other. When the RGB modality only is corrupted, MMTM and MSAF models globally conserve a great performance margin over the unimodal I model without corruption. Indeed, it only happens twice among the 20 V corruptions, with contrast and saturation, that those two multimodal models get under unimodal I (in blue). This leads us to affirm that unimodal corruptions are globally very well handled by the multimodal models that can extract some interesting cues from the corrupted modality while not getting regrettably impacted on the clean modality input most of the time. 

    Comparing the proposed MMSF to the proposed MSAF model, we may observe that MMSF deals better with most corruptions, except for the very challenging ones. Indeed, the I weather alterations are very challenging, and one can see the snow corruption leading, for example, the MSAF model to $81.77$\% mAP, against $46.99$\% mAP for MMSF. In fact, strong corruptions may completely alter $2/3$ of the MMSF fused embedding (Corrupted modality stream feature and the modality shared one), whereas the MSAF attention may simply refactor features in the corrupted modality so that they do not influence too much the final embedding. For weaker corruptions, having a specific stream to mine the right cues while having a specific stream that exploits the correlations among modalities is better. 
    

 \subsubsection{Comparison with state-of-art}\label{sec:SOTA_NCL}

\begin{figure}
    \centering
    \includegraphics[width=
    \linewidth]{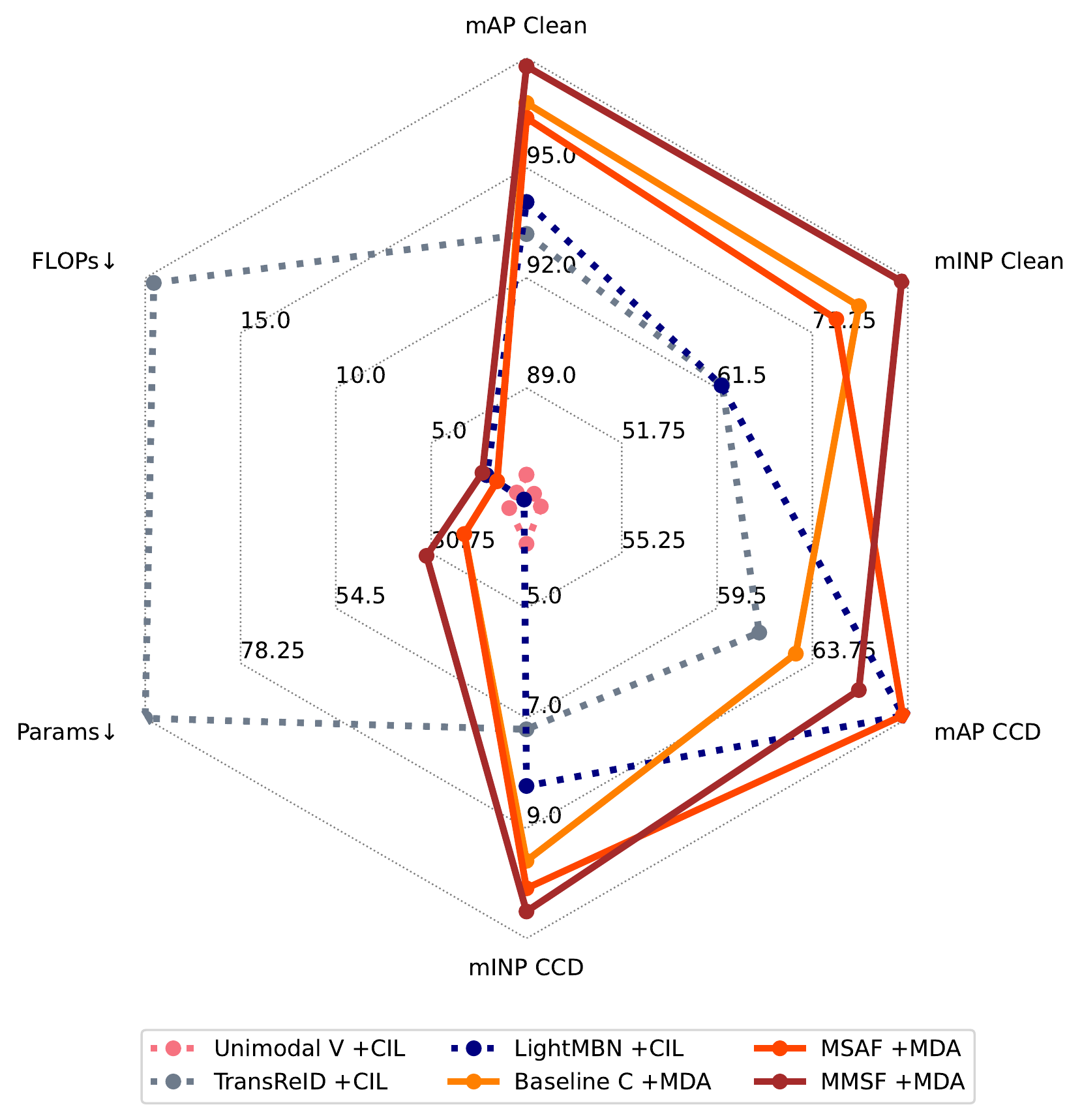}
    \caption{Complexity and accuracy trade-off on the SYSU-MM01 clean and CCD sets. Dashed lines and plain lines are, respectively, unimodal and multimodal approaches. Measures marked with '\textbf{↓}' should be minimized for an optimized model.}
    \label{fig:tradeoff_SYSU}
    \end{figure}

    The multimodal baseline, MSAF, and the proposed MMSF models get compared in terms of complexity and accuracy against the unimodal V and the state-of-art unimodal models LightMBN and TransREID (Fig. \ref{fig:tradeoff_SYSU}). The Appendix 1.D provides detailed performance and additional comparison. 
    Unimodal models are learned using CIL DA, and the multimodal models using our ML-MDA. The accuracy is obtained over both SYSU-MM01 clean and CCD evaluation sets and gathered Fig. \ref{fig:tradeoff_SYSU} along with the models' number of parameters (params) and FLOPs. 

    Performance-wise, each multimodal model is more interesting on clean data than the best unimodal approach, LightMBN. For the best ReID overall, MMSF is the best model, although it performs slightly under MSAF regarding mAP on corrupted data. In fact, MSAF would be favored for a highly challenging environment, especially when facing strong unilateral corruption.

    Complexity-wise, LightMBN is the best model to adopt but comes with a considerable performance decrease from our MMSF, the gap being about $3.32$ mAP PP and $15.59$ mINP PP on clean data.

    \subsubsection{Discussion}\label{sec:Disc_NCL}
    
  Experiments over the SYSU-MM01 dataset give us an excellent overview of the multimodal power under the NCL configuration. The main conclusions are as follows:
   
  \begin{itemize}

      \item The proposed ML-MDA is essential for the multimodal models to handle corruption. This way, models learn how to select the right information from each modality and not get disturbed by noisy features.
      
      \item For the best ReID, the proposed MMSF should be used in priority, followed by MSAF, and finally by the unimodal LightMBN models if the memory resources do not allow it. 
      
      \item The high multimodal accuracy on corrupted data is to highlight as both modalities always get corrupted through the UCD evaluation set, making the task highly challenging.  

      \item The multimodal setting appears as a much better answer to data corruption than the transformer-based approach TransReID, both regarding complexity and accuracy. In fact, TransReID performs less than expected from its performances without DA \citep{chen2021benchmarksDataset_C}, the CIL strategy making, for example, the LightMBN more interesting.
      

            
      
\end{itemize}

\subsection{Scenario with co-located cameras} \label{sec:benchmark_multimodal_CL}


    The spatial alignment brought by co-located V-I cameras should make the correlations from one modality to another easier to find for a model.
    However, this might not make much difference for fusions that come late in the model, as the spatial information will be much diminished and supposedly replaced by semantic information. 
    Also, a corrupted V-I input brings some disequilibrium in how each modality contains relevant information, which should perturb the multimodal models and eventually influence the correlation benefits of spatial alignments. Previous assumptions are explored in the next sections.
    
    

   \subsubsection{Robustness to corruption}\label{sec:Models_corruption_robustness_CL}

    \begin{table*}[htbp]
  \centering
  \caption{Unimodal and multimodal models performances while evaluated on clean and corrupted RegDB datasets.}
    \begin{tabular}{cl||cc|cc|cc|cc}
    \midrule
    
          &    \textbf{Model}   & \multicolumn{2}{c|}{\textbf{Clean}} & \multicolumn{2}{c|}{\textbf{UCD}} & \multicolumn{2}{c|}{\textbf{CCD}} & \multicolumn{2}{c}{\textbf{CCD-50}} \\
          &       & \textbf{mAP} & \textbf{mINP} & \textbf{mAP} & \textbf{mINP} & \textbf{mAP} & \textbf{mINP} & \textbf{mAP} & \textbf{mINP} \\
    \midrule
    \multirow{8}[2]{*}{\begin{sideways}No DA\end{sideways}} & Unimodal V & 99.19 & 96.71 & \textbf{40.54} & \textbf{5.13} & \textbf{40.70} & \textbf{5.01} & \textbf{45.43} & \textbf{6.19} \\
          & Unimodal I & 98.92 & 96.03 & 21.94 & 1.33  & 21.71 & 1.31  & 27.89 & 1.72 \\
          & Baseline S   & 99.39 & 97.60 & 18.66 & 1.75  & 20.73 & 1.57  & 26.86 & 2.17 \\
          & Baseline C & 99.64 & 98.46 & 21.73 & 2.39  & 23.45 & 2.10  & 29.64 & 2.83 \\
          & MAN   & 99.36 & 97.51 & \textcolor[rgb]{ .086,  .086,  .086}{29.02} & \textcolor[rgb]{ .086,  .086,  .086}{3.47} & \textcolor[rgb]{ .086,  .086,  .086}{29.07} & \textcolor[rgb]{ .086,  .086,  .086}{3.15} & \textcolor[rgb]{ .086,  .086,  .086}{35.33} & \textcolor[rgb]{ .086,  .086,  .086}{4.06} \\
          & MMTM  & 99.53 & 98.01 & \textcolor[rgb]{ .086,  .086,  .086}{18.78} & \textcolor[rgb]{ .086,  .086,  .086}{2.06} & \textcolor[rgb]{ .086,  .086,  .086}{19.67} & \textcolor[rgb]{ .086,  .086,  .086}{1.76} & \textcolor[rgb]{ .086,  .086,  .086}{26.15} & \textcolor[rgb]{ .086,  .086,  .086}{2.48} \\
          & MSAF  & \textcolor[rgb]{ .188,  .329,  .588}{99.86} & \textcolor[rgb]{ .188,  .329,  .588}{99.26} & \textcolor[rgb]{ .086,  .086,  .086}{23.42} & \textcolor[rgb]{ .086,  .086,  .086}{2.82} & \textcolor[rgb]{ .086,  .086,  .086}{24.23} & \textcolor[rgb]{ .086,  .086,  .086}{2.37} & \textcolor[rgb]{ .086,  .086,  .086}{31.05} & \textcolor[rgb]{ .086,  .086,  .086}{3.32} \\
          & MMSF  & \textbf{99.88} & \textbf{99.36} & \textcolor[rgb]{ .188,  .329,  .588}{32.63} & \textcolor[rgb]{ .188,  .329,  .588}{5.07} & \textcolor[rgb]{ .188,  .329,  .588}{31.54} & \textcolor[rgb]{ .188,  .329,  .588}{3.79} & \textcolor[rgb]{ .188,  .329,  .588}{38.99} & \textcolor[rgb]{ .188,  .329,  .588}{5.41} \\
    \midrule
    \multicolumn{1}{c}{\multirow{8}[1]{*}{\begin{sideways}ML-MDA / CIL\end{sideways}}} & Unimodal V & 99.51 & 98.21 & 54.61 & 12.58 & 54.61 & 12.51 & 58.43 & 14.56 \\
          & Unimodal T & 98.92 & 96.12 & 44.62 & 6.46  & 44.27 & 6.41  & 50.13 & 8.49 \\
          & Baseline S & 99.87 & 99.37 & 62.48 & 20.34 & 59.33 & 14.60 & 63.89 & 17.34 \\
          & Baseline C  & \textcolor[rgb]{ .188,  .329,  .588}{99.90} & \textcolor[rgb]{ .188,  .329,  .588}{99.45} & 61.92 & 20.14 & 59.06 & 14.64 & 64.15 & 18.08 \\
          & MAN   & \textcolor[rgb]{ .188,  .329,  .588}{99.90} & 99.43 & 62.24 & 23.38 & 60.64 & \textcolor[rgb]{ .188,  .329,  .588}{18.49} & 65.15 & \textcolor[rgb]{ .188,  .329,  .588}{21.62} \\
          & MMTM  & 99.84 & 99.24 & \textcolor[rgb]{ .188,  .329,  .588}{69.06} & \textcolor[rgb]{ .188,  .329,  .588}{25.32} & \textcolor[rgb]{ .188,  .329,  .588}{63.34} & 17.81 & \textcolor[rgb]{ .188,  .329,  .588}{67.27} & 20.17 \\
          & MSAF  & 99.88 & 99.33 & 61.70 & 19.99 & 58.82 & 15.12 & 63.87 & 18.28 \\
          & MMSF  & \textbf{99.95} & \textbf{99.69} & \textbf{76.47} & \textbf{39.51} & \textbf{71.52} & \textbf{30.43} & \textbf{74.25} & \textbf{33.24} \\
    
    \midrule
          
    \end{tabular}%
  \label{tab:unimodal_to_mmodal_perf_analysis_TWORLD_CL} \end{table*}%

    \begin{table*}[htbp]
      \centering
      \caption{Unimodal and multimodal models performances while evaluated on clean and corrupted ThermalWORLD datasets.}
    \begin{tabular}{cl||cc|cc|cc|cc}
    \midrule
    
          &   \textbf{Model}    & \multicolumn{2}{c|}{\textbf{Clean}} & \multicolumn{2}{c|}{\textbf{UCD}} & \multicolumn{2}{c|}{\textbf{CCD}} & \multicolumn{2}{c}{\textbf{CCD-50}} \\
          &       & \textbf{mAP} & \textbf{mINP} & \textbf{mAP} & \textbf{mINP} & \textbf{mAP} & \textbf{mINP} & \textbf{mAP} & \textbf{mINP} \\
    \midrule
    \multirow{8}[2]{*}{\begin{sideways}No DA\end{sideways}} & Unimodal V & 87.38 & 51.71 & 28.74 & 4.50  & 28.97 & 4.47  & 35.28 & 5.18 \\
          & Unimodal I & 56.17 & 10.65 & 24.45 & 3.78  & 24.33 & 3.78  & 27.65 & 3.99 \\
          & Baseline S   & 86.44 & 46.55 & 30.34 & \textcolor[rgb]{ .188,  .329,  .588}{4.84}  & 29.99 & 4.76  & 36.32 & 5.55 \\
          & Baseline C & 87.92 & 50.41 & \textcolor[rgb]{ .188,  .329,  .588}{30.43} & 4.77 & \textcolor[rgb]{ .188,  .329,  .588}{30.51} & \textcolor[rgb]{ .188,  .329,  .588}{4.80} & \textcolor[rgb]{ .188,  .329,  .588}{36.96} & \textcolor[rgb]{ .188,  .329,  .588}{5.65} \\
          & MAN   & 87.50 &  \textcolor[rgb]{ .188,  .329,  .588}{51.98} & 29.10 & 4.56  & 29.15 & 4.54  & 35.62 & 5.26 \\
          & MMTM  & 88.01 & 49.97 & 30.15 & 4.73  & 29.95 & 4.71  & 36.58 & 5.52 \\
          & MSAF  & \textcolor[rgb]{ .188,  .329,  .588}{88.13} & 51.28 & 29.68 & 4.64  & 29.36 & 4.63  & 35.94 & 5.40 \\
          & MMSF  & \textbf{89.43} & \textbf{52.83} & \textbf{30.91} & \textbf{5.20} & \textbf{30.86} & \textbf{5.07} & \textbf{37.44} & \textbf{6.23} \\
    \midrule
    \multicolumn{1}{c}{\multirow{8}[1]{*}{\begin{sideways}ML-MDA / CIL\end{sideways}}} & Unimodal V & 86.37 & 47.42 & 52.77 & 9.51  & 52.83 & 9.43  & 56.28 & 10.79 \\
          & Unimodal I & 55.29 & 9.81  & 32.21 & 4.61  & 32.26 & 4.60  & 34.01 & 4.68 \\
          & Baseline S   & 82.18 & 36.89 & 54.49 & 10.59 & 52.97 & 9.72  & 55.68 & 10.55 \\
          & Baseline C & 86.34 & 43.24 & \textcolor[rgb]{ .086,  .086,  .086}{56.10} & \textcolor[rgb]{ .086,  .086,  .086}{11.04} & 55.20 & 9.93  & 58.01 & 11.02 \\
          & MAN   & 87.11 & 45.47 & 59.22 & 11.19 & 57.54 & 10.56 & 60.23 & 11.64 \\
          & MMTM  & \textbf{87.82} & \textcolor[rgb]{ .188,  .329,  .588}{47.95} & 59.98 & \textcolor[rgb]{ .188,  .329,  .588}{12.55} & \textcolor[rgb]{ .188,  .329,  .588}{58.12} & \textcolor[rgb]{ .188,  .329,  .588}{11.53} & 60.51 & \textcolor[rgb]{ .188,  .329,  .588}{12.36} \\
          & MSAF  & \textcolor[rgb]{ .188,  .329,  .588}{87.62} & \textbf{50.02} & \textcolor[rgb]{ .188,  .329,  .588}{60.38} & \textcolor[rgb]{ .051,  .051,  .051}{11.30} & \textcolor[rgb]{ .051,  .051,  .051}{58.10} & 10.03 & \textcolor[rgb]{ .188,  .329,  .588}{60.78} & \textcolor[rgb]{ .051,  .051,  .051}{10.93} \\
          & MMSF  & 86.10 & 44.50 & \textbf{62.58} & \textbf{14.45} & \textbf{60.75} & \textbf{13.33} & \textbf{62.77} & \textbf{14.24} \\
    \midrule
          
    \end{tabular}%

      \label{tab:unimodal_to_mmodal_perf_analysis_RegDB_CL}%
\end{table*}%

\noindent \textbf{A) Natural models corruption robustness.} 

    To begin with, the RegDB and ThermalWORLD models are learned without the use of data augmentation, and their performances are respectively gathered in the upper half Tab. \ref{tab:unimodal_to_mmodal_perf_analysis_RegDB_CL} and \ref{tab:unimodal_to_mmodal_perf_analysis_TWORLD_CL}. 

    The models must be robust to corrupted data but must also be accurate on clean data at first. Indeed, an optimal model would perform well under the two scenarios. On clean data, the multimodal models are improving over the unimodal visible and Thermal specialists, except for the ThermalWORLD sum model. Precisely, our MMSF model comes first for the two datasets, both regarding mAP and mINP. 
    
    On corrupted evaluation sets, RegDB presents a unimodal visible accuracy considerably ahead of every multimodal model, showing the multimodal model's lack of adaptation while facing corrupted data. Indeed, the unimodal V model is, for example, at $45.43$\% mAP, when the following approach is our MMSF model reaching only $38.99$\% mAP under the CCD-50 set. In reverse, ThermalWORLD observes a considerable improvement with the multimodal setting. Indeed, the unimodal model is behind every multimodal approach for each corrupted dataset version. The most significant improvement comes from the proposed MMSF model again, reaching $40.01$\% mAP, whereas the unimodal V reaches $35.28$\% mAP. 
    
    The lousy thermal modality quality makes the ThermalWORLD dataset distinct from RegDB, which can explain why the multimodal models naturally better handle corruptions. Indeed, this might seem counter-intuitive as a lower quality modality should help less for the ReID, but this more likely indicates that the challenging ThermalWORLD learning environment helps the multimodal models to handle corruption better. This learning configuration forces the models to learn how to adapt regarding each input quality. Under this assumption, higher corruption robustness can be expected from MDA strategies since it works on related concepts by synthetically bringing noisy samples into the learning process. 

    As a supplementary observation, the gap from unimodal to multimodal models performance is much lower under the CL datasets than under SYSU-MM01 and its NCL cameras. Here, the highest performance improvement in PP from the visible to the best multimodal model is about $0,69$ mAP and $2,65$ mINP for RegDB, and about $2,1$ mAP and $1,77$ mINP for ThermalWORLD. In comparison, the gap in PP was about $11.8$ mAP and $39.77$ mINP for SYSU-MM01. This performance gap change might result from the CL cameras concerning RegDB and ThermalWORLD, the additional modality bringing fewer supplementary cues than the NCL setting, as an expected consequence of the spatial alignment. For ThermalWORLD, the gap change is likely also due to the terrible thermal modality quality (BRISQUE value Tab. \ref{tab:datasets_details}), reflected in the mAP gap from the unimodal V to the unimodal T model, being of $31.21$ PP. For RegDB, the unique camera per modality probably influence this aspect as well, making the problem easier, leading to almost maxed-out performances that do not allow similar improvement through the multimodal setting.

 \noindent \textbf{B) DA impact on models robustness.} 

    Models performances while considering data augmentation strategies are presented lower half Tab. \ref{tab:unimodal_to_mmodal_perf_analysis_RegDB_CL} and \ref{tab:unimodal_to_mmodal_perf_analysis_TWORLD_CL} respectively for RegDB and ThermalWORLD. 
    
    Moving from no use of DA to its usage leads to impressive performance improvements on corrupted data. Where the RegDB multimodal models performed lower than the unimodal visible model using no DA, all multimodal models learned with our ML-MDA become way ahead of the visible model. The greatest improvement comes from unimodal V to our proposed MMSF model, which increases the mAP by $16.91$ PP for CCD and $15.82$ PP for CCD-50. 
    
    For corrupted versions of ThermalWORLD, for which multimodal models already had better performances than unimodal specialists before DA, the performance gap significantly increases with ML-MDA usage. Considering CCD evaluation, for example, the gap from unimodal V to the best approach being MMSF is about $7.92$ mAP, where it was about $1.89$ mAP percentage points without DA. 
    
    The massive multimodal corruption robustness improvement from the proposed multimodal data augmentation on the two datasets makes it a crucial approach. With it, the MMSF model becomes the best working approach for RegDB, followed by MMTM. In fact, modalities are both corrupted most of the time, so the attention through MMTM and MSAF probably becomes tough to adjust for the models. MMSF does not allow another modality to bring additional noise in its modality-specific streams and consequently better benefits from each input. Also, its central stream can focus only on the encoding of the modality correlations and eventually improve the ReID even more.

        \begin{figure*}
     \centering
     \begin{subfigure}[b]{0.49\textwidth}
         \centering
         \includegraphics[width=\textwidth]{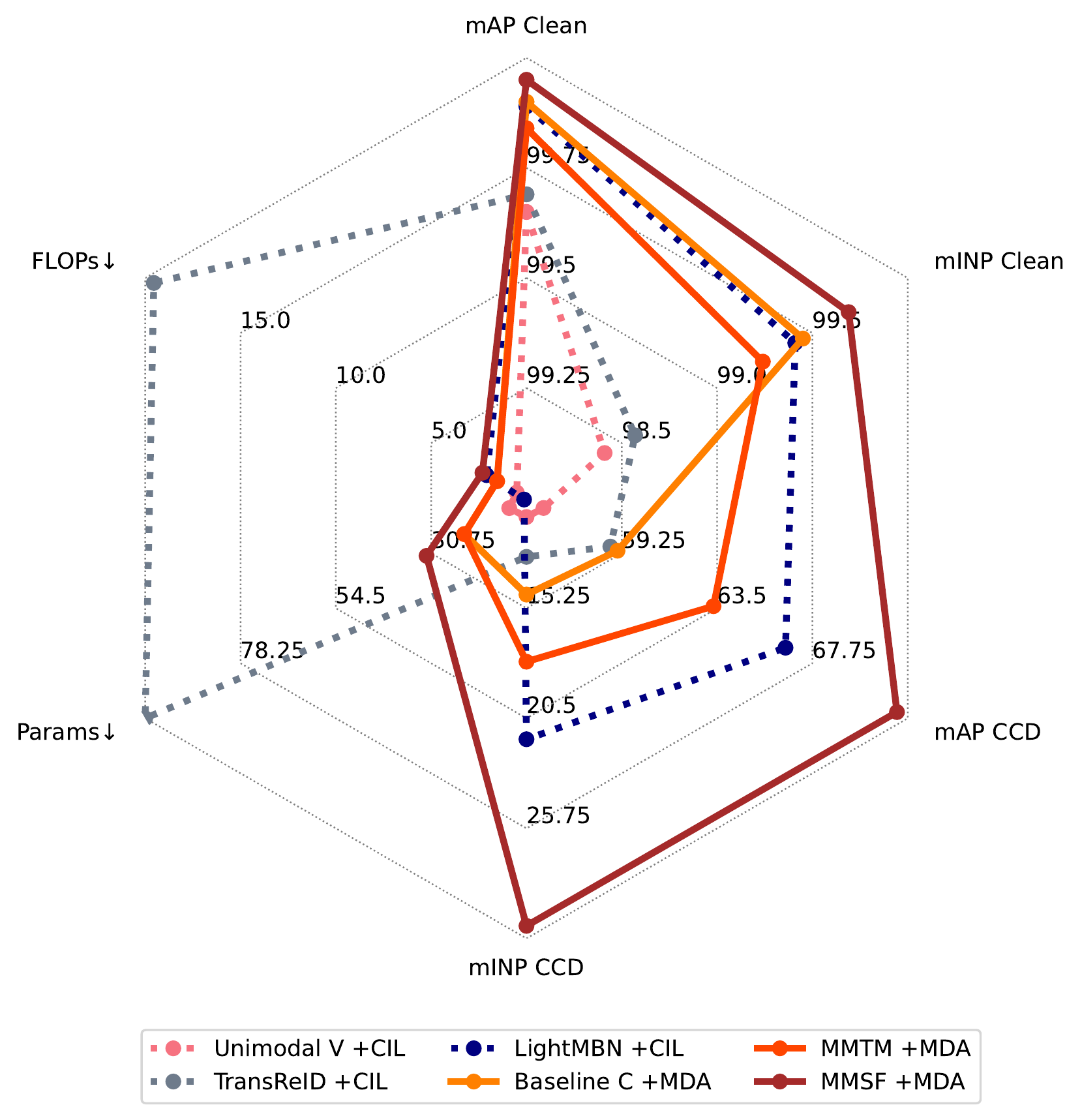}
         \caption{RegDB}
         \label{fig:RegDB_tradeoff}
     \end{subfigure}
     \hfill
     \begin{subfigure}[b]{0.49\textwidth}
         \centering
         \includegraphics[width=\textwidth]{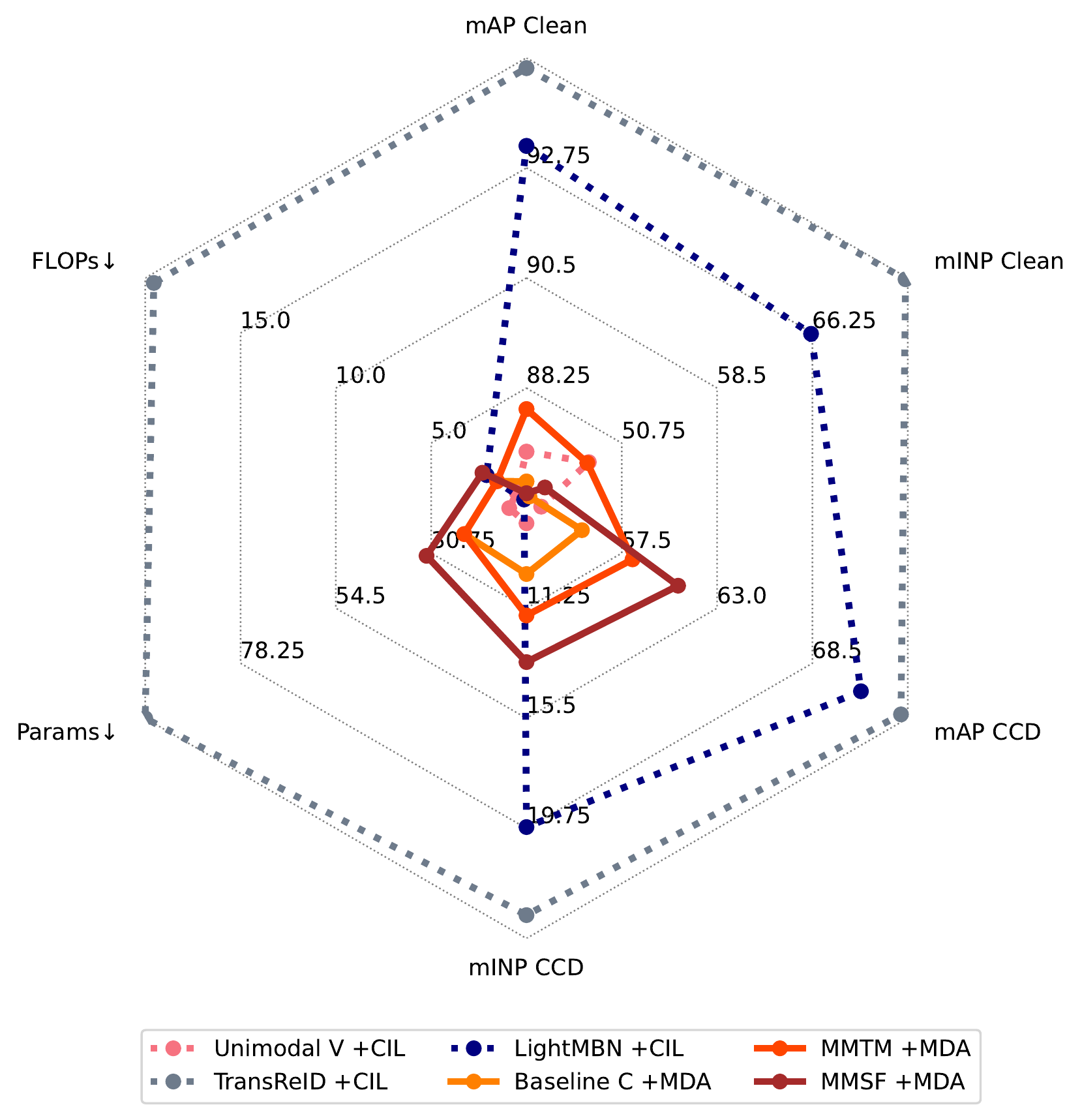}
         
         \caption{ThermalWORLD}
         \label{fig:TWORLD_tradeoff}
     \end{subfigure}
     \caption{Complexity and accuracy trade-off using clean and CCD evaluation sets. Dashed lines and plain lines are, respectively, unimodal and multimodal approaches. Measures marked with '\textbf{↓}' should be minimized for an optimized model.}
     \label{fig:CL_tradeoff}
    \end{figure*}
    
    

    \subsubsection{Comparison with state-of-the-art}\label{sec:SOTA_CL} 

    For CL cameras, multimodal MMSF and MMTM models get compared to the state-of-art unimodal models under both RegDB and ThermalWORLD Clean and CCD evaluation sets. The accuracy is put in perspective of the models' complexity through their number of parameters (params) and FLOPs (Fig. \ref{fig:CL_tradeoff}). Only the CCD evaluation set is considered as this configuration is the most adapted to CL cameras (Section \ref{sec:UCD_CCD}) and should allow drawing the main conclusions.
    
    For RegDB (Fig. \ref{fig:RegDB_tradeoff}), the best-performing model is our proposed MMSF model in terms of accuracy, both on clean and corrupted data. The model is followed by LightMBN and then by MMTM. Hence, the complexity and accuracy trade-off comes between LightMBN and the MMSF model, MMSF being the best way for a strong ReID, and LightMBN for a lighter but lesser efficient approach. 
    
    
    Focusing on ThermalWORLD (Fig. \ref{fig:TWORLD_tradeoff}), the story is different. Despite the same CL camera configuration as RegDB, the two compared multimodal models are much less accurate than the unimodal LightMBN and TransReID models while having more parameters and needing more FLOPs than LightMBN. This large gap in behavior from RegDB to ThermalWORLD comes from the latter dataset's infrared quality again. Still, for a similar ResNet-18 backbone architecture through Unimodal V, we observe that the multimodal models are more accurate. This shows how the multimodal models can benefit from the additional modality even if this one is of low quality, but that it is not enough to compare with LightMBN and TransReID discriminant power.
    Finally, among TransReID and LightMBN models, it is again an accuracy and complexity trade-off. Heavier but more discriminant is TransReID for ThermalWORLD, and much lighter but also less discriminant is LightMBN. 
    
        

\subsubsection{Discussion}

    The previous analysis under the CL setting from RegDB and ThermalWORLD datasets allowed us to reinforce some observations from the NCL setting and draw additional conclusions that are as follows:
    
    \begin{itemize}

    \item The proposed ML-MDA data augmentation is crucial for a multimodal model to handle challenging data in NCL and CL settings well. Also, models still benefit much from the MDA when the original dataset is challenging, as observed through ThermalWORLD. 
    
    \item Our MMSF model deals substantially better with clean and corrupted data than every other approach, including TransReID, despite its highlighted corruption dealing \citep{chen2021benchmarksDataset_C}. The early fusion likely allows the model to better apprehend and disentangle the corrupted features from the clean ones between modalities. Considered attention approaches exchange stream information later in the process and consequently have already lost an essential part of the modality correlations. Plus, they do not have modality-specific streams as MMSF, whereas it assures that the final embedding conserves features from a good modality definition and also ensure the model does not only focus on modality-shared features.


    \item The deficient infrared data quality of the ThermalWORLD dataset does not allow the multimodal setting to compare with unimodal state-of-art.
    
    
    
    \end{itemize}

\section{Conclusion}

Real-world surveillance and especially person ReID is a complex task that requires models to handle complex and abstract concepts, handle data corruption and remain lightweight. To address these challenges, the multimodal setting can be a powerful tool, as an additional modality brings supplementary information that can help to reach higher accuracy while it allows reaching competitive complexity thanks to lightened backbones. However, real-world conditions and the subsequent data corruptions (e.g., weather, blur, illumination) have to be considered. To this aim, our study proposes a strong V-I multimodal evaluation through the first V-I corrupted evaluation sets (UCD and CCD) for multimodal (and cross-modal) V-I person ReID, tackling the lack of multimodal real-world datasets \citep{rahate2022multimodal}. Precisely, 20 visible and 19 infrared corruptions are considered, 3 datasets, 2 camera settings (NCL and CL), 2 state-of-art person ReID models, a MDA, 6 multimodal models, comprising 3 attention-based, 2 baselines, and our proposed MMSF architecture.

Experiments on the clean and proposed corrupted datasets converge to present the proposed ML-MDA as a must-use to make any multimodal model way more robust to real-world events. The multimodal models observe a larger margin of improvement from the NCL rather than the CL scenario as a consequence of the additional information provided by the NCL complementary view. Still, the benefits of plural modalities are unequivocal for both scenarios, the TransReID model being way more complex and less accurate than plural multimodal approaches (except under really low-quality infrared through the ThermalWORLD dataset). Especially, among multimodal approaches, our MMSF model comes ahead of every considered model for the two scenarios, highlighting the importance of considering modality-specific features not tackled in attention SOA models.  

To extend this work, vision-based MDA could be further explored as it showed great benefits but remains not much investigated in the literature. Also, the proposed MMSF has shown weakness while facing strongly and unilaterally corrupted data, which has less impact on attention-based models. Hence, adding the right attention modules may allow getting the best of both worlds. Finally, different backbones could be explored for a better accuracy/complexity ratio. 


\section*{Declarations}


\begin{itemize}
\item \textbf{Acknowledgements:} This research was supported by Nuvoola AI Inc., the Natural Sciences and Engineering Research Council of Canada, and by Compute Canada (www.computecanada.ca).
\item \textbf{Availability of data and materials.} 

\textbf{SYSU-MM01.} A signed dataset release \href{agreement}{https://github.com/wuancong/SYSU-MM01/blob/master/agreement/agreement.pdf} must be sent to wuancong@gmail.com and wuanc@mail.sysu.edu.cn to obtain a download link. 

\textbf{RegDB.} The dataset can be downloaded by submitting a copyright form this \href{website}{http://dm.dongguk.edu/link.html}. 

\textbf{ThermalWORLD.} A part of the dataset is available only. It can be downloaded from this \href{link}{https://drive.google.com/file/d/1XIc_i3mp4xFlDJ_S5WJYMJAHq107irPI/view}, obtained from GitHub ThermalGAN \href{issues}{https://github.com/vlkniaz/ThermalGAN/issues/12}.
\item \textbf{Code availability.}
GitHub 
 \href{link}{https://github.com/art2611/MREiD-UCD-CCD.git}.
\end{itemize}







\bibliography{sn-bibliography}

\newpage

\begin{appendices}

    \section{Details regarding infrared corruptions}

\begin{figure*}
    
    \centering
    \includegraphics[width=\textwidth,height=8cm]{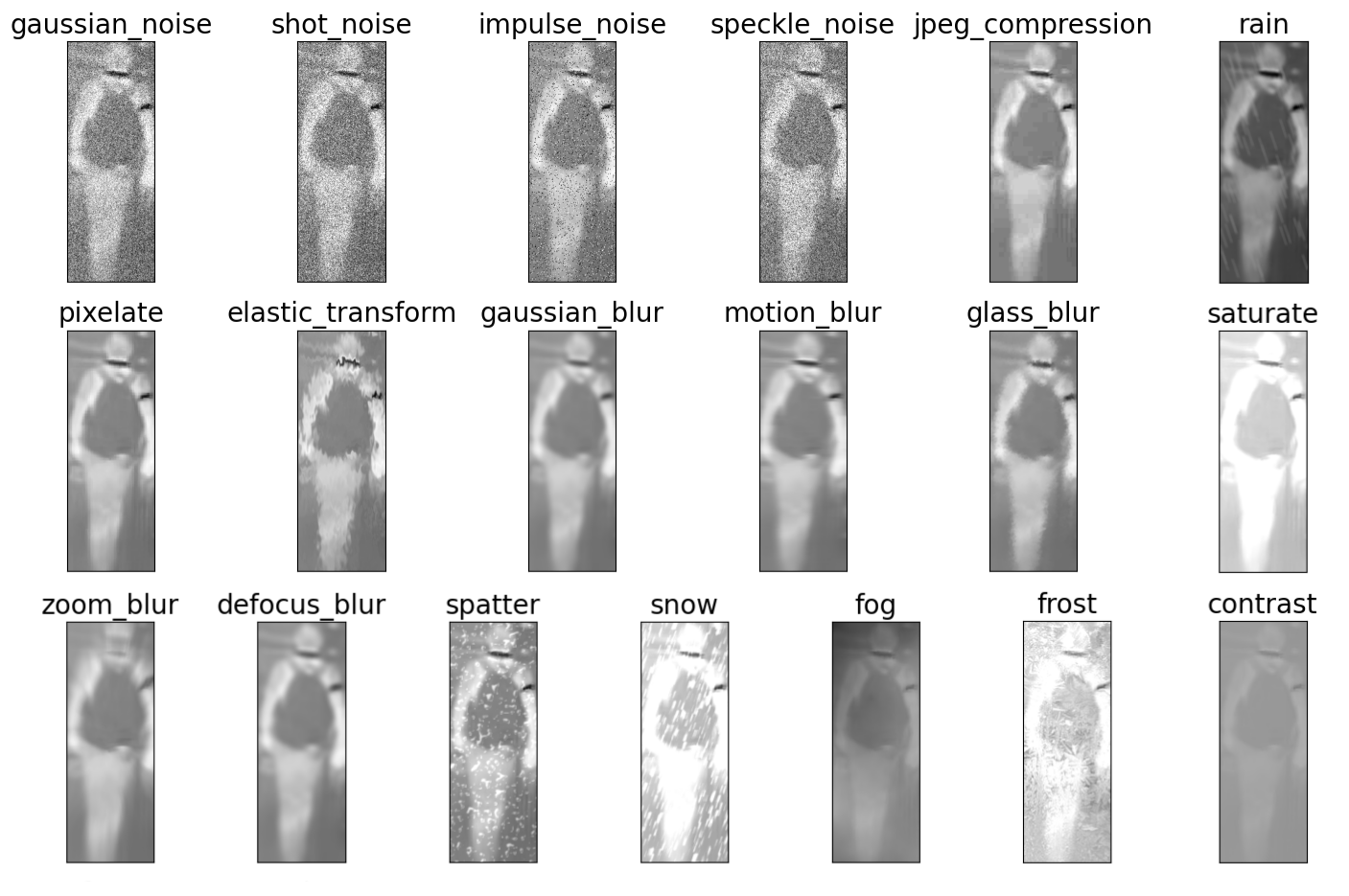}
    \caption[width=0.6\textwidth]{Taxonomy of the 19 thermal corruptions, all applied with an intensity level 3.}
    \label{fig:thermal_corruptions_level3}

    \end{figure*}

    Further details are provided Tab. \ref{tab:RGB_to_IR_corrupt} concerning the way infrared corruptions were obtained from the existing visible ones. Also, a figure gathering an example of each 19 infrared corruptions is presented Fig. \ref{fig:thermal_corruptions_level3}.

        \begin{table}[htbp]
      \centering
      \caption{Applied corruption adjustments to extend Visible (V) corruptions to the Infrared (I) modality. V corruptions that get grayscaled to perform I corruptions appear in red.}
    \begin{tabular}{c|l|l}
    \toprule
    Type & V corruption & \multicolumn{1}{l}{I corruption} \\
    \midrule
    \multirow{4}[2]{*}{\begin{sideways}Noise\end{sideways}} & \textcolor{red}{Gaussian noise} & \multicolumn{1}{l}{\multirow{4}{12em}{Each noise is used similarly but is first grayscaled.}} \\
          & \textcolor{red}{Shot noise} &  \\
          & \textcolor{red}{Impulse noise} &  \\
          & \textcolor{red}{Speckle noise}  &  \\
    \midrule
    \multirow{5}[2]{*}{\begin{sideways}Blur\end{sideways}} & Defocus blur & \multicolumn{1}{l}{\multirow{5}{12em}{No change in the way blurs are extended to infrared.}} \\
          & Glass blur  &  \\
          & Motion blur  &  \\
          & Zoom blur  &  \\
          & Gaussian blur &  \\
    \midrule
    \multirow{6}[2]{*}{\begin{sideways}Weather\end{sideways}} & Snow  & \multicolumn{1}{l}{\multirow{6}{12em}{Brightness is not used for Infrared. Spatter (water or dirt splash) and frost get grayscaled. Others are similarly applied.}} \\
          & \textcolor{red}{Frost} &  \\
          & Fog   &  \\
          & Rain  &  \\
          & Brightness &  \\
          & \textcolor{red}{Spatter}  &  \\
    \midrule
        
        \multirow{5}[2]{*}{\begin{sideways}Digital\end{sideways}} & Contrast & \multicolumn{1}{l}{\multirow{5}{12em}{Digital corruptions are the same except for saturation. Saturation for infrared make close objects brighter.}} \\
          & Elastic trsf &  \\
          & Pixelate  &  \\
          & JPEG compr &  \\
          & Saturation  &  \\
    \bottomrule
    \end{tabular}%

      \label{tab:RGB_to_IR_corrupt}%
    \end{table}%


    \section{MMSF optimization}

    \subsection{MMSF and not co-located cameras}
    
        The MMSF model fuses the features from each visible and infrared backbone in its middle stream (Sec. 3.2 in main document). The proper fusion location has to be determined. The fusion can either be early in the process, fusing directly original images by element-wise sum, or later by fusing feature maps from each modality stream the same way for a given layer. Intuitively, as the cameras are not co-located and, as a consequence, the images not spatially aligned for the SYSU dataset, an early fusion in the middle stream might result in a noised fused representation. Indeed, the model in early stages might not be able to extract meaningful representation and adapt them according to the used fusion. In reverse, later-stage feature maps have a superior degree of abstraction and should suit better such fusion. Also, considering a corrupted evaluation setting, corruptions may increase the representation gap from one modality to another and thus eventually make the model further benefit from a later fusion. Still, as earlier representation gather more information and hence more potential correlations from one modality to another, one can only be assured of where to fuse data in the middle stream with an empirical study.

        Obtained results are gathered in Tab. \ref{tab:MMSF_fuse_loc_NC}. As expected, fusing at later stages for the middle stream leads to a more discriminant final representation. Indeed, performances in both mAP and mINP gradually improve from fusing at $k=0$ to fusing at $k=4$ for clean data. For example, the mAP improves by $1,18$\% and the mINP by $7,27$\% from $k=0$ to $k=4$ respectively. Also, for the UCD corrupted setting, mAP improves by $2,09$\% and $0,24$\% mINP for the same k values. Later fusion is more beneficial to the model, confirming the drawn hypothesis on NCL data. Most complex cases are similarly handled by all configurations on corrupted data according to the mINP witch evolves from $10.27$\% to $10.51$\% mINP for $k=0$ to $k=4$ respectively. . 
        
        \begin{table}[htbp]
          \centering
          \caption{MMSF performances regarding the fusion location in the middle stream, and for the clean and UCD SYSU datasets.}
            \begin{tabular}{l||cc|cc}
            \midrule
            \textbf{Model}      & \multicolumn{2}{c|}{\textbf{SYSU}} & \multicolumn{2}{c}{\textbf{SYSU-UCD}} \\
                  & \textbf{mAP} & \textbf{mINP} & \textbf{mAP} & \textbf{mINP} \\
            \midrule
            MMSF 0 & 96.59 & 73.11 & 63.73 & 10.27 \\
            MMSF 1 & 97.27 & 77.01 & 64.28 & 9.57 \\
            MMSF 2 & 97.28 & 77.37 & 63.43 & 9.60 \\
            MMSF 3 & 97.76 & 79.91 & 64.81 & 10.38 \\
            MMSF 4 & \textbf{97.77} & \textbf{80.38} & \textbf{65.82} & \textbf{10.51} \\
            \midrule
            \end{tabular}%
          \label{tab:MMSF_fuse_loc_NC}%
        \end{table}%

    \subsection{MMSF and co-located cameras}
   
    The MMSF model under CL cameras may behave differently than under NCL cameras due to the alignment of the visible and thermal images in a given pair. In fact, earlier fusion should allow more correlation findings as the feature representation is less compressed than later in the process. Plus, the spatial alignment should make the sum of the feature maps relevant even in the early process. Still, an earlier MMSF fusion comes with a more complex architecture since it requires more layers in the central stream, which needs to be kept in mind. 

    Performances regarding the fusion location l for RegDB and ThermalWORLD datasets are presented Tab. \ref{tab:MMSF_fuse_loc_CL}. As expected, it is interesting to observe the performance decrease from $\textit{l}=0$ to $\textit{l}=4$ on RegDB for clean and CCD-50 data. In practice, the mAP decreases by $2.46$\%, and the mINP by $3.74$\% on CCD-50 dataset. ThermalWORLD results are not following this same scheme, as the results on clean data are the highest for $\textit{l}=1$, followed by $\textit{l}=2$ $\textit{l}=4$ and $\textit{l}=0$. The model act as an in-between the NCL and CL settings, which probably comes from the thermal modality being of terrible quality, messing with the expected impact of spatial alignment. Still, the RegBD model acts similarly as the ThermalWORLD one on corrupted data, performing the best through earlier fusions. In fact, earlier fusion may allow the model to get less impacted by corrupted features as the model can directly find and discard them while benefiting from the most correlations. 
    
    \begin{table}[htbp]
      \centering
      \caption{MMSF performances regarding the fusion location in the middle stream, and for the clean and CCD-50 versions of RegDB and ThermalWORLD datasets.}
        \begin{tabular}{cl||cc|cc}
        \midrule
              &  \textbf{Model}     & \multicolumn{2}{c|}{\textbf{Clean}} & \multicolumn{2}{c}{\textbf{CCD-50}} \\
              &       & \textbf{mAP} & \textbf{mINP} & \textbf{mAP} & \textbf{mINP} \\
        \midrule
        \multirow{5}[2]{*}{\begin{sideways}RegDB\end{sideways}} & MMSF 0 & \textbf{99.95} & \textbf{99.69} & \textbf{74.25} & \textbf{33.24} \\
              & MMSF 1 & 99.93 & 99.60 & 73.16 & 30.36 \\
              & MMSF 2 & 99.93 & 99.66 & 73.17 & 30.68 \\
              & MMSF 3 & 99.93 & 99.64 & 72.55 & 29.74 \\
              & MMSF 4 & 99.94 & 99.64 & 71.79 & 29.50 \\
        \midrule
        \multicolumn{1}{c}{\multirow{5}[1]{*}{\begin{sideways}TWORLD\end{sideways}}} & MMSF 0 & 86.10 & 44.50 & \textbf{62.77} & \textbf{14.24} \\
              & MMSF 1 & \textbf{86.27} & \textbf{45.96} & 62.27 & 13.59 \\
              & MMSF 2 & 86.28 & 45.26 & 62.06 & 13.44 \\
              & MMSF 3 & 86.14 & 44.24 & 61.21 & 12.98 \\
              & MMSF 4 & 86.58 & 44.89 & 61.30 & 12.95 \\
        \midrule
              
        \end{tabular}%
      \label{tab:MMSF_fuse_loc_CL}%
    \end{table}%

    \section{Element-wise sum or concatenation}

    \subsection{Fusion with co-located cameras}\label{sec:data_fusion_NCL}

    The absence of spatial alignment may favor the concatenation over the element-wise sum of the feature vectors or vice-versa. Indeed, a summation might require the vector information to be aligned from one modality to another, not to erase it. Unlike element-wise sum, concatenation conserves features from each modality the same way and could consequently better fit with NCL configuration. Also, in the case of corrupted data, corruption should also tend to make the produced modality-specific feature vector representation different and hence favored concatenation as well. Still, this is only hypothetical as the information may be only semantic and aligned at this point of the data encoding. 

    To confirm or invalidate the previous assumptions, the baseline, MMTM, and MSAF models are compared in terms of mAP and mINP regarding a sum or a concatenation of the feature vectors, and on clean and UCD SYSU-MM01 datasets (Table \ref{tab:sum_or_cat_NC}). The corrupted UCD set is only considered here as uncorrelated corruptions are the most suited for the NCL configuration and as it should allow answering the previous hypothesis. Models were trained using ML-MDA, but the ML-MDA is beyond this section's scope, only used as a tool (for now) to bring consistency from clean to corrupted evaluation. Observing the baseline results, it seems beneficial to concatenate the features as expected while looking at clean data results. Indeed, concatenation improves mINP by $1.52$\% while conserving similar mAPs. However, performances under the UCD dataset show that summing is more beneficial, slightly improving the mAP and mINP respectively by $0.60$ and $0,21$\%. These results on corrupted data are going against our hypothesis, as concatenation was expected to overpass summation under UCD. Observing attention MMTM and MSAF models results on clean and UCD data; the former considerably improves from summation to concatenation, whereas the second considerably decreases. Hence, the concatenation or summation of the features at such a level of abstraction probably allows the model to deal with the absence of spacial alignment and to align features according to the fusion used. Consequently, the best feature vector fusion strategy is model dependent and needs to be assessed experimentally. For MMTM and MSAF, the upcoming NCL analysis will consider only their best fusion version MMTM C and MSAF S.

    \begin{table}[htbp]
      \centering
      \caption{Baseline, MMTM, and MSAF performances on SYSU-MM01 dataset while summing (S) or concatenating (C) their feature vectors. Clean and UCD evaluation only are considered since UCD respects the most NCL corruptions (Section 4.3 main document).}
        \begin{tabular}{l||cc|cc}
        \midrule
        \textbf{Model}     & \multicolumn{2}{c|}{\textbf{Clean}} & \multicolumn{2}{c}{\textbf{UCD}} \\
              & \textbf{mAP} & \textbf{mINP} & \textbf{mAP} & \textbf{mINP} \\
        \midrule
        Baseline S & 96.54 & 74.49 & \textbf{64.00} & \multicolumn{1}{c}{\textbf{9.72}} \\
        Baseline C & \textbf{96.77} & \textbf{76.01} & 63.40 & \multicolumn{1}{c}{9.51} \\
        \midrule
        MMTM S & 94.97 & 68.33 & 63.29 & 9.45 \\
        MMTM C & \textbf{95.81} & \textbf{74.23} & \textbf{64.41} & \textbf{11.49} \\
        \midrule
        MSAF S & \textbf{96.36} & \textbf{73.70} & \textbf{67.78} & \textbf{10.09} \\
        MSAF C & 96.04 & 71.13 & 66.20 & 9.68 \\
        \midrule
        \end{tabular}%
      \label{tab:sum_or_cat_NC}%
    \end{table}%

    \subsection{Fusion with co-located cameras}\label{sec:data_fusion_CL}
    
    The best strategy between element-wise sum and concatenation of the feature maps was shown to be model-dependent for NCL cameras (Section \ref{sec:data_fusion_NCL}). Unlike NCL cameras, CL ones bring spacial alignment that might impact the preferred fusion differently. An empirical analysis is provided Tab. \ref{tab:sum_or_cat_CL} to determine which fusion to follow and if it remains model dependent by applying it on the baseline, MMTM and MSAF models. 
    In practice, where it behaves similarly for each dataset by favoring the fusion by concatenation for the baseline models, it becomes more complex for MMTM and MSAF models. Indeed, the MMTM and the MSAF models, which exchange information between the visible and thermal CNN streams, seem not to follow a specific rule again. More than being model-dependent, performances appear as being data-dependent. For example, MMTM S performs better under both clean and corrupted RegDB settings, whereas it is MMTM C for ThermalWORLD. It is important to notice that the performance gap can be important from sum to concatenation, making such analysis important while seeking the right way to fuse feature vectors in a model. Models performing best for MMTM and MSAF are kept for the rest of CL cameras study. \\ 
    
    \begin{table}[htbp]
      \centering
      \caption{Baseline, MMTM and MSAF perfor-
mances on RegDB and ThermalWORLD datasets while summing (S) or concatenating (C)
the feature vectors as fusion.}
    \begin{tabular}{cl||cc|cc}
    \midrule
    
          &    \textbf{Model}   & \multicolumn{2}{c|}{\textbf{Clean}} & \multicolumn{2}{c}{\textbf{CCD-50}} \\
          &       & \textbf{mAP} & \textbf{mINP} & \textbf{mAP} & \textbf{mINP} \\
    \midrule
    \multirow{6}[6]{*}{\begin{sideways}RegDB\end{sideways}} & Baseline S & 99.87 & 99.37 & 63.89 & 17.34 \\
          & Baseline C & \textbf{99.90} & \textbf{99.45} & \textbf{64.15} & \textbf{18.08} \\
\cmidrule{2-6}          & MMTM S & \textbf{99.84} & \textbf{99.24} & \textbf{67.27} & \textbf{20.17} \\
          & MMTM C & 99.80 & 99.12 & 63.92 & 17.97 \\
\cmidrule{2-6}          & MSAF S & 99.84 & 99.19 & 59.22 & 13.26 \\
          & MSAF C & \textbf{99.88} & \textbf{99.33} & \textbf{63.87} & \textbf{18.28} \\
    \midrule
    \multicolumn{1}{c}{\multirow{6}[5]{*}{\begin{sideways}TWORLD\end{sideways}}} & Baseline S & 82.18 & 36.89 & 55.68 & 10.55 \\
          & Baseline C & \textbf{86.34} & \textbf{43.24} & \textbf{58.01} & \textbf{11.02} \\
\cmidrule{2-6}          & MMTM S & 86.17 & 45.50 & 59.60 & 10.91 \\
          & MMTM C & \textbf{87.82} & \textbf{47.95} & \textbf{60.51} & \textbf{12.36} \\
\cmidrule{2-6}          & MSAF S & \textbf{87.62} & \textbf{50.02} & 60.78 & 10.93 \\
          & MSAF C & 87.73 & 48.00 & \textbf{60.57} & \textbf{12.01} \\
    \midrule
          
    \end{tabular}%
      \label{tab:sum_or_cat_CL}%
    \end{table}%

    \section{Detailed complexity and accuracy trade-off}
    
    The accuracy and complexity analysis is provided in the main document Section 6.2.3 for NCL and 6.3.2 for CL cameras. However, detailed performances and complexity was not provided. Hence, this section focus on the detailed models performances for NCL and CL cameras at first and finally present the complexity in terms of parameters and FLOPs for each and every considered model.  
    
    \subsection{Accuracy with not co-located cameras}
     
    State-of-art unimodal models, along with the unimodal V model, are compared to the baseline C, MMSF, and MSAF multimodal approaches learned using our ML-MDA (Table \ref{tab:SOTA_NC}). Unimodal models are evaluated while being learned with and without the CIL strategy. As a first observation, multimodal models are all considerably over the unimodal models in terms of both mAP and mINP on clean data. Indeed, the highest improvement in mAP and mINP  from the best unimodal model performances is respectively about $3.32$\% and $15.59$\%.  Then, if we compare the multimodal models among themselves, MMSF comes first by improving mAP of the baseline by $1.00$\% while it improves its mINP by $4.37$\%. Surprisingly, MSAF is below the baseline's mINP by $2.31$\% while conserving its mAP. \\

    Looking now at the UCD performances, the best working model is MSAF with $67.78$\% mAP and $10.09$\% mINP. Then, LightMBN and MMSF are pretty equivalent, with respectively mAPs about $67.80$\% and $65.82$\% but mINPs about $8.23$\% and $10.51$\%. From the previous observations, both the MSAF and MMSF models can be used to improve over the state-of-art unimodal models, considering both clean and corrupted data. However, the benefits from the proposed MMSF are higher than the ones from the MSAF approach. Still, if the real-world conditions were expected as tough, MSAF would eventually be favored. However, if conditions were varying or tending to be clean, MMSF should be used. \\
    
    \begin{table}[htbp]
      \centering
          \caption{Multimodal comparison with the state-of-art unimodal models. MDA refer to our ML-MDA approach. }
        \begin{tabular}{cl||cc|cc}
            \midrule
        
              &       & \multicolumn{4}{c}{\textbf{SYSU}} \\
              &   \textbf{Model}    & \multicolumn{2}{c|}{\textbf{Clean}} & \multicolumn{2}{c}{\textbf{UCD}} \\
              &       & \textbf{mAP} & \textbf{mINP} & \textbf{mAP} & \textbf{mINP} \\
        \midrule
        \multirow{3}[2]{*}{\begin{sideways}No DA\end{sideways}} & Unimodal V & 86.25 & 39.97 & 32.36 & 1.91 \\
              & TransReID & 94.33 & 64.79 & 52.03 & 3.60 \\
              & LightMBN  & 94.45 & 64.06 & 40.90 & 2.13 \\
        \midrule
        \multirow{3}[2]{*}{\begin{sideways}CIL\end{sideways}} & Unimodal V  & 86.64 & 42.78 & 51.64 & 3.83 \\
              & TransReID  & 93.20 & 62.02 & 61.38 & 7.20 \\
              & LightMBN  & 94.07 & 61.95 & \textcolor[rgb]{ .188,  .329,  .588}{67.80} & 8.23 \\
        \midrule
        \multirow{3}[1]{*}{\begin{sideways} MDA\end{sideways}} & Baseline C & \textcolor[rgb]{ .188,  .329,  .588}{96.77} & \textcolor[rgb]{ .188,  .329,  .588}{76.01} & 63.01 & 9.59 \\
              & MSAF  & 96.36 & 73.70 & \textbf{67.78} & \textcolor[rgb]{ .188,  .329,  .588}{10.09} \\
              & MMSF  & \textbf{97.77} & \textbf{80.38} & 65.82 & \textbf{10.51} \\
            \midrule
        \end{tabular}%
      \label{tab:SOTA_NC}%
    \end{table}%
    
    \subsection{Accuracy with co-located cameras}
     
    The multimodal models trained using our ML-MDA are compared with state-of-art unimodal frameworks learned using CIL DA under the CL setting Table \ref{tab:SOTA_comparison_CL}. First, observing performances on RegDB clean data, the multimodal baseline C and the proposed MMSF are ahead of the unimodal models. MMSF improving LightMBN mAP and mINP respectively from $99.89$ to $99.92$ and $99.45$ to $99.57$. If we observe corrupted performances, only the proposed MMSF can improve over the best unimodal model LightMBN + CIL, increasing the mAP by $4.97\%$ on CCD and by $4.85$ on CCD-50. Hence, our MMSF model is the way to go for both clean and corrupted data under the CL configuration performance-wise. 
    
    About ThermalWORLD, models behave really differently. The TransReID and LightMBN models perform much better than the best multimodal approach MSAF on clean data. Indeed, for example, TransReID reaches $95.86\%$ mAP when MSAF reaches $87.82\%$ mAP. In fact, the slight $0.87\%$ mAP improvement from the Unimodal V to the MSAF model shows how hard the multimodal setting benefits from the bad thermal modality. This is confirmed by the results under corrupted settings, as the best multimodal approach MMSF is $12.86\%$ and $13.44$\% mAP below the TransReID approach for CCD and CCD-50 respectively. Consequently, favoring stronger unimodal models is a better strategy when the supplementary modality is far behind in terms of quality. 

\begin{table*}[htbp]
   \centering
    \caption{RegDB and ThermalWORLD - Comparison with SOTA}
    \begin{tabular}{cl||cc|cc|cc}
    \toprule
          &   \textbf{Model}    & \multicolumn{2}{c|}{\textbf{Clean}} & \multicolumn{2}{c|}{\textbf{CCD}} & \multicolumn{2}{c}{\textbf{CCD-50}} \\
          &       & \textbf{mAP} & \textbf{mINP} & \textbf{mAP} & \textbf{mINP} & \textbf{mAP} & \textbf{mINP} \\
    \midrule
    \multirow{9}[2]{*}{\begin{sideways}RegDB\end{sideways}} & Unimodal V & 99.26 & 96.64 & 45.15 & 7.01  & 45.42 & 6.20 \\
          & TransReID & 99.34 & 97.35 & 45.64 & 5.69  & 48.60 & 7.01 \\
          & LightMBN  & \textcolor[rgb]{ .051,  .051,  .051}{99.90} & \textcolor[rgb]{ .051,  .051,  .051}{99.41} & 32.40 & 7.01  & 33.63 & 3.85 \\
          & Unimodal V + CIL & 99.65 & 98.41 & 55.76 & 10.9  & 58.53 & 14.8 \\
          & TransReID + CIL & 99.69 & 98.57 & 58.74 & 12.8  & 60.48 & 16.2 \\
          & LightMBN + CIL & \textcolor[rgb]{ .051,  .051,  .051}{99.89} & \textcolor[rgb]{ .051,  .051,  .051}{99.41} & \textcolor[rgb]{ .188,  .329,  .588}{66.55} & \textcolor[rgb]{ .188,  .329,  .588}{21.5} & \textcolor[rgb]{ .188,  .329,  .588}{69.40} & \textcolor[rgb]{ .188,  .329,  .588}{26.2} \\
          & Baseline C + ML-MDA & \textcolor[rgb]{ .188,  .329,  .588}{99.90} & \textcolor[rgb]{ .188,  .329,  .588}{99.45} & 59.06 & 14.6  & 64.15 & 18.0 \\
          & MMTM + ML-MDA & 99.84 & 99.2!4 &  63.34 & 17.8  & 67.27 & 20.1 \\
          & MMSF + ML-MDA & \textbf{99.95} & \textbf{99.69} & \textbf{71.52} & \textbf{30.4} & \textbf{74.25} & \textbf{33.2} \\
    \midrule
    \multirow{9}[1]{*}{\begin{sideways}ThermalWORLD\end{sideways}} & Unimodal V & 86.44 & 49.44 & 28.06 & 3.86  & 35.27 & 5.18 \\
          & TransReID & \textbf{95.86} & \textbf{77.98} & 65.47 & 17.2  & 68.66 & 20.0 \\
          & LightMBN  & 93.02 & 65.94 & 37.34 & 5.60  & 44.01 & 6.70 \\
          & Unimodal V + CIL & 86.95 & 48.07 & 52.85 & 7.97  & 56.33 & 10.6 \\
          & TransReID + CIL & \textcolor[rgb]{ .188,  .329,  .588}{94.79} & \textcolor[rgb]{ .188,  .329,  .588}{73.82} & \textbf{73.61} & \textbf{23.1} & \textbf{76.21} & \textbf{25.8} \\
          & LightMBN + CIL & 93.20 & 66.14 & \textcolor[rgb]{ .188,  .329,  .588}{71.30} & \textcolor[rgb]{ .188,  .329,  .588}{19.7} & \textcolor[rgb]{ .188,  .329,  .588}{73.62} & \textcolor[rgb]{ .188,  .329,  .588}{21.4} \\
          & Baseline C + ML-MDA & \textcolor[rgb]{ .086,  .086,  .086}{86.34} & \textcolor[rgb]{ .086,  .086,  .086}{43.24} & \textcolor[rgb]{ .086,  .086,  .086}{56.10} & \textcolor[rgb]{ .086,  .086,  .086}{9.93} & \textcolor[rgb]{ .086,  .086,  .086}{58.01} & \textcolor[rgb]{ .086,  .086,  .086}{11.02} \\
          
          & MMTM + ML-MDA & \textcolor[rgb]{ .086,  .086,  .086}{87.82} & \textcolor[rgb]{ .086,  .086,  .086}{47.95} & \textcolor[rgb]{ .086,  .086,  .086}{58.12} & \textcolor[rgb]{ .086,  .086,  .086}{11.53} & \textcolor[rgb]{ .086,  .086,  .086}{60.51} & \textcolor[rgb]{ .086,  .086,  .086}{12.36} \\
          & MMSF + ML-MDA & \textcolor[rgb]{ .086,  .086,  .086}{86.10} & \textcolor[rgb]{ .086,  .086,  .086}{44.50} & \textcolor[rgb]{ .086,  .086,  .086}{60.75} & \textcolor[rgb]{ .086,  .086,  .086}{13.33} & \textcolor[rgb]{ .086,  .086,  .086}{62.77} & \textcolor[rgb]{ .086,  .086,  .086}{14.24} \\
        \bottomrule
    \end{tabular}%
  \label{tab:SOTA_comparison_CL}%
\end{table*}%

     \subsection{Models complexity}


    Thanks to the additional modality and knowledge, a multimodal setting might allow the use of lighter backbones than a given unimodal pipeline while matching or even improving accuracy. From the previous experiments, the multimodal accuracy comes ahead unimodal approaches, but a complexity analysis remains needed, and is provided Tab. \ref{tab:complexity_analysis}. The analysis is presented regarding the models' number of parameters and the FLOPs needed to compute a single input. 

    First, one can observe that the TransReID complexity appeals at first sight, much heavier through $102.0$M parameters than any other models, followed by MMSF with $\textit{l}=0$ and its $34.6$M parameters. The lighter model is LightMBN, being more than ten times lighter than TransReID with $7.6$M parameters. Based on the obtained results for NCL, the multimodal setting improves much the ReID accuracy. Especially, LightMBN comes first among unimodal approaches but is way less performing than MSAF and MMSF. In practice, the proposed MMSF works the best ($\textit{l}=4$ for NCL) under NCL cameras and should be used if resources allow it, requiring $2.31$ GFLOPs and $31.9$M parameters. Otherwise, MSAF would be the next model to go with $1.54$ GFLOPs and $22.5$M parameters, finally followed by the unimodal LightMBN approach with $2.09$GFLOPs and $7.6$M parameters. 
    
    Considering the CL setting, the proposed MMSF ($\textit{l}=0$) model is ahead, followed directly by the LightMBN model performance-wise. Similarly LightMBN comes with less complexity than MMSF, thus making a compromise between precision and complexity.

    \begin{table}[htbp]
      \centering
      \caption{Size (Number of parameters) and computation complexity regarding FLOPs. }
        \begin{tabular}{ll||cc}
        \toprule
        \multicolumn{2}{c|}{\textbf{Model}} & \textbf{No params (M)} & \textbf{FLOPs (G)} \\
        \midrule
        \multicolumn{2}{l|}{Unimodal V or I} & 11.3 & 0.51 \\
        \multicolumn{2}{l|}{TransReID} & 102.0 & 19.55 \\
        \multicolumn{2}{l|}{LightMBN} & 7.6  & 2.09 \\
        \midrule
        \multicolumn{2}{l|}{Baseline} & 22.5 & 1.54 \\
        \multicolumn{2}{l|}{MAN} & 22.5 & 1.54 \\
        \multicolumn{2}{l|}{MMTM} & 23.8 & 1.54 \\
        \multicolumn{2}{l|}{MSAF} & 22.5 & 1.54 \\
        \multicolumn{2}{l|}{MMSF \textit{l}=0} & 34.6 & 3.09 \\
        \multicolumn{2}{l|}{MMSF \textit{l}=4} & 31.9 & 2.31 \\
        \bottomrule
        \end{tabular}%
      \label{tab:complexity_analysis}%
    \end{table}%

    \section{Qualitative analysis }

    Models learned through ML-MDA were compared over clean and corrupted data in terms of performances Section 6.2.1 in the main document. However, observing what the models are focusing on to discriminate and ReID would be a great way to draw additional conclusions, or at least to better understand why a model is better than another. To this end, adapted for pairwise matching algorithms, similarity based Class Activation Maps (CAMs) from \cite{simcamstylianou2019visualizing} is used. It is important to notice that MMSF CAMs are produced from its two modality specific streams only, and that the shared modality stream cannot be analysed from this CAM technique for the NCL cameras. Indeed, CAMs could be determined for the middle stream but there would be no way to dissociate from which spatial part of the V or I modality comes the shared activation.
    
\begin{figure*}
    \centering
    \includegraphics[width=0.7
    \textwidth]{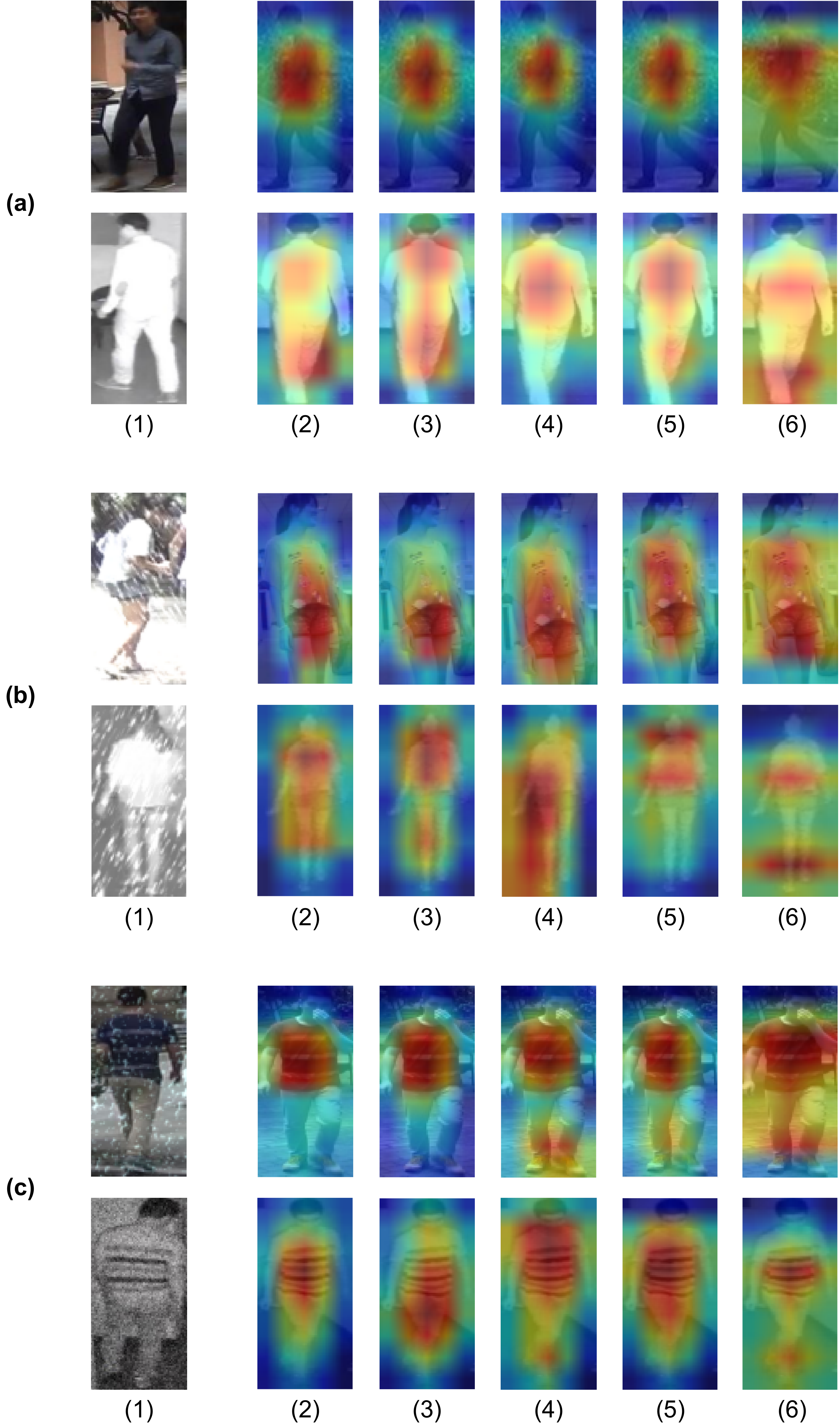}
    \caption[width=0.6\textwidth]{Three examples of similarity based CAMs using SYSU-MM01 (a) clean V-I pairs, (b) snow corrupted V-I pairs, (c) differently corrupted V (spatter) and I (gaussian noise) pairs. CAMs are computed from (2) baseline C, (3) MAN, (4) MMTM, (5) MSAF, and (6) MMSF. (1) is the reference V-I pair.}
    \label{fig:CAM}
    \end{figure*}

    To put visualizations in perspective, models ranking performance-wise on clean data start from MMSF, followed by MAN, Baseline C, MSAF, and MMTM. Observing Fig. \ref{fig:CAM}a., one can see that activation on the V modality is more or less similar from one model to another, focusing mainly on the torso. Actually, MMSF might appear a bit less accurate but tend to focus on the same region. However, it seems that the most discriminant models consider both the torso and the legs of the person concerning the I modality. Indeed, from MMSF (6) to MAN (3), Baseline C (2), MSAF (5) and finally to MMTM (4), legs activation just decreases.
    
    Switching to corrupted data, models ranking was the following under CCD: MSAF, MMSF, MMTM, Baseline C, MAN. Looking at Fig. \ref{fig:CAM}b. one may observe that the best working models focus both on the short and on the t-shirt of the individual concerning the V modality. About the I modality, it is harder to interpret, as the added snow made the focus of the models much less accurate, which in fact correlate well with the snow corruption impact on the thermal modality (Tab. \ref{tab:individual_corruption_impact}). In fact, for I, both MMSF and MSAF focus on waist, but MMSF adding feet where MSAF adds shoulders to it. Also, MMTM seems much perturbed, as its attention is not so much on the person, and baseline C with MAN are both looking pretty fuzzy, mainly looking at the whole back of the individual. 
    
    If we look at corruptions that seem to less affect each modalities, with Fig. \ref{fig:CAM}c., one can see that the thermal modality (gaussian noise) is much better apprehended by each model. For the most discriminant ones, MMSF (6), MSAF (5) and MMTM (4), it is interesting to again observe the importance of feet in the ReID process. 

\end{appendices}





\end{document}